\documentclass{article}

\PassOptionsToPackage{numbers,compress}{natbib}
\usepackage[main,final]{neurips_2026}

\usepackage[utf8]{inputenc}
\usepackage[T1]{fontenc}
\usepackage{hyperref}
\usepackage{url}
\usepackage{booktabs}
\usepackage{amsfonts}
\usepackage{nicefrac}
\usepackage{microtype}
\usepackage{xcolor}
\usepackage{enumitem}
\usepackage{wrapfig}
\usepackage{adjustbox}

\usepackage{pgfplots}
\pgfplotsset{compat=1.18}
\usepackage{graphicx}
\usepackage{tikz}
\usetikzlibrary{positioning,calc}
\usepackage{makecell}
\usepackage{subcaption}
\usepackage{tabularx,array}
\usepackage{amsmath,amssymb,mathtools}
\usepackage{placeins}
\usepackage{wrapfig}
\usepackage{tabularx}
\usepackage{booktabs}

\newcommand{\R}{\mathbb{R}}          
\newcommand{\SE}{\mathrm{SE}}        
\newcommand{\Normal}{\mathcal{N}}
\newcommand{\Ad}{\mathrm{Ad}}
\newcommand{\ad}{\mathrm{ad}}

\newcommand{\mypara}[1]{\medskip\noindent\textbf{#1}~}

\begin{document}


\title{Change-Robust Online Topological Memory for Long-Term Relocalization and Semantic Navigation}

\author{%
  Jiaming Wang\thanks{Equal contribution.} \And
  Jizhuo Chen\footnotemark[1] \And
  Diwen Liu\footnotemark[1] \And
  Atharva Ghotavadekar \AND
  Jiaxuan Da \And
  Linh K\"astner \And
  Harold Soh \AND
  National University of Singapore
}

\maketitle

\begin{abstract}
Long-term semantic navigation requires a robot to reuse past observations after
appearance and scene change, but semantic memories are only useful if the robot
can relocalize into the memory without corrupting it with false visual matches.
We propose CROSS, a change-robust topological memory that introduces a
\emph{pre-commitment} localization layer between visual place recognition and
map update.  Instead of treating a retrieved keyframe as an immediate place
association or loop-closure factor, CROSS lifts each RGB-D retrieval into a
candidate global $\mathrm{SE}(3)$ pose mode using relative pose estimation.
A bounded Gaussian-mixture filter then propagates competing continuous
trajectory branches with odometry, rejects branches that are physically
inconsistent, and promotes only persistent branches to loop closures.  This
moves ambiguity handling from discrete place IDs or post-hoc graph-factor
rejection to continuous pose-space validation before map commitment.  Across
public long-term relocalization benchmarks and real quadruped object-navigation
experiments, CROSS improves reuse of a single sparse RGB-D memory under
illumination, seasonal, dynamic-scene, and object-level change.
Project page: \url{https://jiaming.im/CROSS/}
\end{abstract}

\section{Introduction}
\label{subsec:introduction}

Robots operating over long time horizons need spatial memories that remain useful after illumination, seasonal, object-level, and dynamic-scene changes. Such memories should support semantic navigation tasks such as object search or language-goal navigation, but these tasks first require reliable spatial reuse: the robot must know where relevant observations were made, how to reach them, and where it is relative to them.

A common approach attaches object labels, language embeddings, or scene-graph structure to a SLAM-built metric map~\cite{chaplot2020object, chang2024goat, gu2024conceptgraphs, jatavallabhula2023conceptfusion, wang2024probable}. This is effective when the metric substrate remains valid, but long-term deployment exposes a brittle dependency: if relocalization fails under appearance change or stale map priors, the semantic layer becomes difficult to use.

We instead study a sparse topological memory for long-term semantic navigation. Each node stores a raw RGB-D observation and a pose distribution, while edges encode local navigability. Keeping raw observations is deliberately task-agnostic: the robot need not know during mapping which object categories or language goals will be queried later. Stored observations can be processed at query time by detectors~\cite{zhou2022detecting, minderer2022simple}, VLMs~\cite{bai2025qwen3}, or other visual foundation models~\cite{radford2021learning}.

The central localization challenge is deciding \emph{when to trust} a visually plausible association. Modern visual place recognition (VPR) improves recall under appearance change~\cite{ali2024boq, berton2025megaloc}, yet high-scoring retrievals can still be false positives in repetitive or changed environments. Topological and topometric methods reduce such errors by filtering over place IDs, route positions, or image sequences~\cite{cummins2008fabmap2,labbe2013appearance,maddern2011catslam,vysotska2016lazy,xu2020probvpr,xu2021topometric,suomela2024placenav}. These methods delay commitment, but usually over discrete states, discarding much of the relative-pose geometry needed to test physical compatibility with odometry. Metric SLAM back-ends preserve geometry, but candidate loop closures typically enter the graph as factors before robust mechanisms accept or reject them~\cite{sunderhauf2012switchable,olson2013inference,latif2013robust,hsiao2019mhisam2}. This leaves a missing interface: visual associations should remain uncommitted, yet still be tested as continuous pose evidence.

\begin{figure}
    \centering
    \includegraphics[width=1\linewidth]{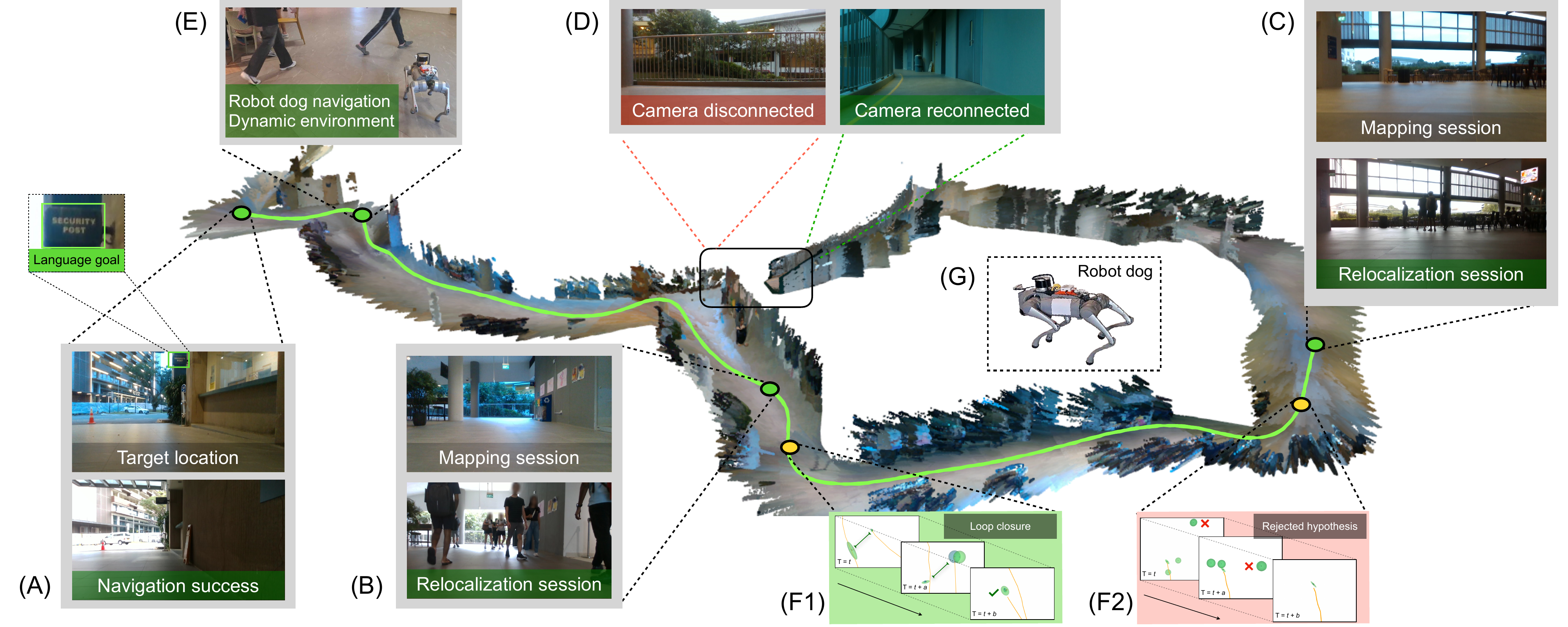}
    \caption{\small{CROSS supports long-term semantic navigation with a sparse RGB-D topological memory under lighting variation, object rearrangement, dynamic pedestrians, and sensor degradation. Visual retrievals are lifted into continuous $\mathrm{SE}(3)$ hypotheses, temporally validated by odometry, and only persistent branches are promoted to loop closures or relocalization events.}}
    \label{fig:teaser}
\end{figure}

CROSS (\textbf{C}hange-\textbf{R}obust \textbf{O}nline \textbf{S}patial memory \textbf{S}ystem) addresses this interface with a pre-commitment localization layer. VPR proposes candidate keyframes; RGB-D relative pose estimation lifts each candidate into a global $\mathrm{SE}(3)$ pose mode. A bounded Gaussian-mixture belief then propagates competing continuous trajectory branches with odometry, prunes motion-inconsistent branches, and promotes only persistent branches to loop closures, tracking recovery, or kidnapped-robot relocalization. The contribution is not any individual component in isolation, but the commitment boundary: ambiguous visual associations are validated in continuous pose space before they modify the topological memory.

This paper makes three contributions:
\begin{itemize}[leftmargin=15pt,itemsep=1pt, topsep=0pt, parsep=0pt, partopsep=0pt]
  \item An online topological memory that stores raw RGB-D keyframes as task-agnostic semantic anchors while maintaining pose uncertainty for navigation. Code is available at \url{https://github.com/jiaming-ai/CROSS}.
  \item A retrieval-to-pose hypothesis layer that converts visual matches into continuous $\mathrm{SE}(3)$ trajectory branches and validates them against odometry before map commitment.
  \item An evaluation on public long-term relocalization benchmarks and real-robot object-goal navigation, showing improved memory reuse under illumination, seasonal, dynamic-scene, and object-level change.
\end{itemize}

\vspace{-0.5em}
\section{Related Work}
\label{sec:related_work}

\vspace{-0.5em}
\paragraph{Semantic maps and long-term robot memory.}
Many semantic navigation systems attach object, language, or scene-graph
information to a metric map built by SLAM, including 2D semantic occupancy
maps~\cite{chaplot2020object,chang2024goat,liu2024okrobot,wang2024probable},
3D scene graphs~\cite{rosinol2022hydra,Werby-RSS-24,gu2024conceptgraphs}, and
dense open-vocabulary feature maps~\cite{jatavallabhula2023conceptfusion,
zhu2024sni,li2024sgs,zhou2024feature}.  These representations provide rich
semantic interfaces, but their spatial reuse depends on reliable relocalization
in the underlying metric map.  CROSS is complementary: it does not aim to be a
richer semantic representation, but a change-robust spatial memory substrate.
By storing raw RGB-D keyframes, it defers semantic interpretation to query time
while focusing on whether past observations can be localized and reused after
appearance or scene change.

\vspace{-0.5em}
\paragraph{Topological, topometric, and route-based localization.}
Appearance-based systems such as FAB-MAP and RTAB-Map perform probabilistic
place recognition and online map growth in the space of visual
observations~\cite{cummins2008fabmap2,labbe2013appearance}.  CAT-SLAM,
SeqSLAM-style sequence matching, visual teach-and-repeat, and
experience-based navigation add temporal or route consistency to improve
long-term operation under changing appearance~\cite{maddern2011catslam,
milford2008mapping,vysotska2016lazy,furgale2010visual, churchill2013experience}.  More recent topometric systems combine VPR with
Bayesian filtering, odometry-conditioned transitions, and off-map
states~\cite{xu2020probvpr,xu2021topometric,suomela2024placenav}.  These
methods are robust and scalable because their beliefs are tied to place IDs,
route positions, or particles over a fixed topometric state space.  CROSS uses
the same insight---visual matches should be temporally validated---but changes
the state being validated: visually retrieved keyframes are lifted into
continuous $\mathrm{SE}(3)$ pose modes, so the filter can test full geometric
compatibility with odometry before committing to a place association.

\vspace{-0.5em}
\paragraph{Multi-hypothesis localization and robust SLAM.}
Multi-modal pose beliefs have long been studied in Markov and Monte Carlo
localization~\cite{fox1999mcl,thrun2001robust}, and modern multi-hypothesis
SLAM back-ends represent ambiguous odometry, data association, or loop closures
as multiple factor-graph solutions~\cite{hsiao2019mhisam2,hsiao2020aras}.
Robust pose-graph methods such as switchable constraints, max-mixtures, and
related loop-closure consistency checks downweight or reject erroneous loop
closures after candidate factors have been proposed~\cite{sunderhauf2012switchable,
olson2013inference,latif2013robust}.  CROSS is not a replacement for these
back-ends.  Its contribution is a front-end, pre-commitment layer for visual
topological memory: candidate visual associations remain as bounded continuous
pose branches, are propagated and pruned by sequential motion consistency, and
are promoted to loop-closure factors only after sustained support.

\begin{figure}[t]
    \centering
    \includegraphics[width=0.99\linewidth]{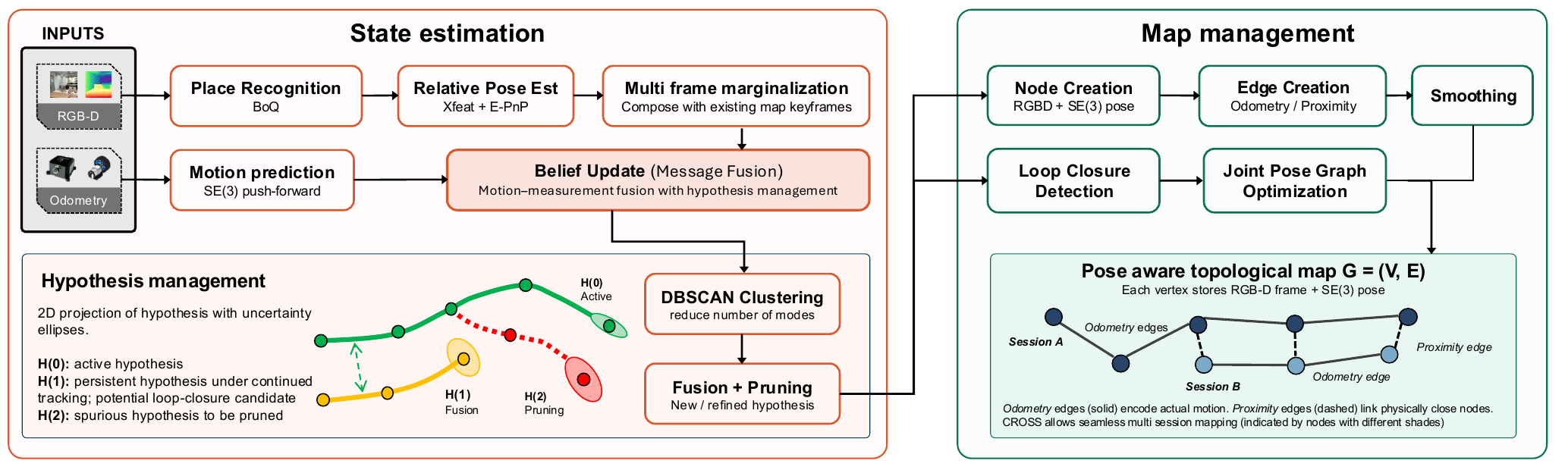}
    \caption{\small{\textbf{System Overview.}
    Given an RGB-D frame and odometry, the online tracking module (orange) performs temporal hypothesis filtering in continuous $\mathrm{SE}(3)$. Motion updates are propagated via $\mathrm{SE}(3)$ push-forward, while measurement updates are constructed through VPR-based keyframe retrieval. Competing hypotheses are efficiently managed using Gaussian-mixture clustering, pruning, and fusion (see Section~\ref{sec:state-estimation}). 
    The map-management module (blue) uses the estimated state from the tracking module to create nodes and edges, detect loop closures, and perform pose-graph optimization and smoothing, yielding a consistent pose-aware topological map (see Section~\ref{sec:map-management}).}}
    \label{fig:system-overview}
    \vspace{-1em}
\end{figure}

\vspace{-0.5em}
\section{Problem Formulation}
\label{sec:problem-formulation}
\vspace{-1em}

We aim to construct an environment representation that supports
robust navigation in real-world settings subject to natural and
often substantial change. Such a representation should satisfy three
desiderata: (i) \emph{VLM compatibility}, enabling semantic inference directly
over stored observations; (ii) \emph{change robustness}, allowing the robot to
localize and reuse a single map despite significant appearance variation
without remapping; and (iii) \emph{local navigability}, providing sufficient
spatial structure to support short-horizon pose estimation and goal-directed
motion.

Guided by these requirements, we represent the environment as a sparse,
pose-aware topological graph $\mathcal{G}=(\mathcal{V},\mathcal{E})$.
Each node $v_i\in\mathcal{V}$ stores an RGBD keyframe $z_i$ and an associated
camera pose $c_i\in\mathrm{SE}(3)$.
The robot pose at time $t$ is represented by a random variable
$x_t\in\mathrm{SE}(3)$, and the robot state is summarized as $(x_t,n_t)$, where
$n_t$ indexes the current (or nearest) keyframe.
Importantly, both $x_t$ and $\{c_i\}$ are treated as random variables and are modeled using finite Gaussian mixtures on $\mathrm{SE}(3)$ (Section~\ref{sec:state-estimation}), which enables a compact representation of multi-modal pose uncertainty.

We use $u_t$ to denote the odometry increment from $x_{t-1}$ to $x_t$.
Given observations $z_{1:t}$ and odometry inputs $u_{1:t}$, our goal is to
maintain the joint posterior over the robot trajectory and the topological map,
$
p(x_{1:t},\,\mathcal{G} \mid z_{1:t},\,u_{1:t}).
$
The current node index $n_t$ is inferred from a
pose estimate $\hat{x}_t$ as
$
n_t = \arg\min_{v_i \in \mathcal{V}} d(\hat{x}_t, \hat{c}_i),
$
where $d(\cdot)$ denotes a pose distance on $\mathrm{SE}(3)$ and $\hat{c}_i$
denotes a representative estimate of the keyframe pose (e.g., the highest-weight
mixture component).

This formulation is topological in what it stores, but continuous in what it
tests.  The graph contains sparse RGB-D keyframes rather than persistent
landmarks, voxels, or fused semantic embeddings used in metric SLAM and
metric-semantic maps~\cite{campos2021orb,labbe2019rtab,
jatavallabhula2023conceptfusion,gu2024conceptgraphs}.  However, each visual
association is evaluated as a constraint on the robot's continuous pose.  Thus,
the latent ambiguity is not only ``which node is this?'' but ``which
$\mathrm{SE}(3)$ trajectory branch makes this visual association physically
consistent with the odometry?''  This distinguishes CROSS from topological,
topometric, and route-based localization methods whose beliefs are tied to
place IDs or route states~\cite{cummins2008fabmap2,maddern2011catslam,
furgale2010visual,churchill2013experience,xu2021topometric,suomela2024placenav},
and from multi-hypothesis or robust pose-graph methods that reason about
ambiguous associations after they have been introduced as candidate
factors~\cite{hsiao2019mhisam2,sunderhauf2012switchable,olson2013inference,
latif2013robust}.

\section{Method: CROSS}
\vspace{-0.5em}
In this section, we describe CROSS (overview in Fig.~\ref{fig:system-overview}). At a high level, the system is organized into state estimation and map management. Direct inference over the joint trajectory--map posterior is intractable because trajectory and map estimates are coupled, so we use the conditional factorization
\begin{equation}
p(x_{1:t}, \mathcal{G} \mid z_{1:t}, u_{1:t})
=
\underbrace{
p\!\left(\mathcal{G} \mid x_{1:t}, z_{1:t}, u_{1:t}\right)
}_{\text{map management}}
\cdot
\underbrace{
p\!\left(x_{1:t} \mid z_{1:t}, u_{1:t}\right)
}_{\text{state estimation}} .
\label{eq:factorization}
\end{equation}
Due to space constraints, we focus on the key ideas in each module and relegate details to Appendix~\ref{app:method-details}. 

\vspace{-0.5em}
\subsection{State Estimation: Approximate Inference and Hypothesis Management}
\label{sec:state-estimation}
To represent multi-modal pose uncertainty while keeping computation bounded, we model both the robot pose $x_t$ and keyframe poses $\{c_i\}$ as finite Gaussian mixtures on $\mathrm{SE}(3)$. Since $\mathrm{SE}(3)$ is a nonlinear manifold, we define each Gaussian in the local tangent space $\mathfrak{se}(3)$ of its mean using the logarithm map, and reconstruct poses via the exponential map, 
\begin{equation}
\small
p(s_j) \approx \sum_{k=1}^{K} w_j^{(k)}
\mathcal{N}_{\mathfrak{se}(3)}
\!\left(\log\!\big((\mu_j^{(k)})^{-1}s_j\big);\mathbf{0},\Sigma_j^{(k)}\right),
\quad s_j \in \{x_t,c_i\}.
\label{eq:se3-gmm}
\end{equation}

where each component specifies a local distribution around its mean on $\mathrm{SE}(3)$:
$
x_t^{(k)} = \mu_t^{(k)}\exp(\xi),\;
 \!c_i^{{(k)}} = \mu_i^{(k)}\exp(\xi),\;
 \!\xi \sim \mathcal{N}(0,\Sigma^{(k)}).
$
The residual form $\log\!\big((\mu)^{-1}(\cdot)\big)$ ensures that uncertainty is expressed in the tangent space where linearization is valid. 

We perform inference over the current pose by forward message passing on the factor graph (Fig. \ref{fig:pgm-factor-graph} in Appendix) induced by the motion and measurement factors. Let $m_t^{\text{mot}}(x_t)$ and $m_t^{\text{meas}}(x_t)$ denote the motion and
measurement messages arriving at node $x_t$. The belief is given by:
\begin{equation}
\small
p(x_t \mid z_{1:t},u_{1:t},\mathcal{G})\propto
\underbrace{m_t^{\text{meas}}(x_t)}_{\text{measurement message}}
\underbrace{m_t^{\text{mot}}(x_t)}_{\text{motion message}} .
\label{eq:filter-master}
\end{equation}
Eq.~\eqref{eq:filter-master} is evaluated in closed-form using the approximations below and is similar to the classic Gaussian-sum filter (GSF)~\cite{ristic2003beyond, alspach2003nonlinear}, but differs in how the {measurement} message is constructed.

\vspace{-0.5em}
\mypara{Forward motion message.}
The motion factor is
$
\phi_t^{\text{mot}}(x_{t-1},x_t,u_{t})
\triangleq
p(x_t \mid x_{t-1},u_{t}) ,
$
and its forward message to $x_t$ is the Bayes prediction
$
m_t^{\text{mot}}(x_t)
=
\int
p(x_t \mid x_{t-1},u_{t})\,
p(x_{t-1}\mid z_{1:t-1},u_{1:t-1})\,dx_{t-1}.
$
Let
$\{w^{(k)}_{t-1},\mu^{(k)}_{t-1},\Sigma^{(k)}_{t-1}\}_{k=1}^K$ denote the
weights, means, and covariances of the mixture at time $t{-}1$. We approximate the pushed-forward
mixture at time $t$ as
$
\small
m_t^{\text{mot}}(x_t)
\approx
\sum_{k=1}^K
w^{(k)}_{t-1}\,
\mathcal{N}_{\mathfrak{se}(3)}
\!\left(
    \log\!\left( (\mu^{(k)}_{t\mid t-1})^{-1} x_t \right);
    \; 0,\Sigma^{(k)}_{t\mid t-1}
\right),
$
where each mixand is mapped to a new Gaussian with mean
$\mu^{(k)}_{t\mid t-1}= \mu^{(k)}_{t-1}\,u_t$ 
and covariance
$
\Sigma^{(k)}_{t\mid t-1}
\;\approx\;
\mathrm{Ad}_{u_t^{-1}}\,
\Sigma^{(k)}_{t-1}\,
\mathrm{Ad}_{u_t^{-1}}^{\!\top}
\;+\;
Q_t .
$
Here, $u_t$ is the odometry input, and $Q_t$ is the process-noise covariance in $\mathfrak{se}(3)$.
The mixture weights are preserved
($w^{(k)}_{t\mid t-1}=w^{(k)}_{t-1}$) because the motion kernel is probability-preserving (see Appendix~\ref{appx:prediction})

\vspace{-0.5em}
\mypara{Measurement Message.}
A standard fixed-support Gaussian-sum filter refines only the pose modes already
present in the prior (see Appendix~\ref{app:gsf}). Long-term relocalization, however, requires a measurement
to introduce modes at arbitrary previously mapped locations.  We therefore
construct a global measurement message whose components are born from retrieved
keyframes rather than from the current dominant tracking branch.
To overcome this limitation, we introduce a \emph{global measurement} message. 
We begin with a local measurement model between the current RGB-D observation $z_t$ and node $i$. The relative pose can be estimated
using standard PnP methods~\cite{lepetit2009ep,quan1999linear,urban2016mlpnp}:
$
T_{i,t}
\;=\;
f_{\mathrm{PnP}}(z_t, z_i),
$
where $z_t$ and $z_i$ denote the RGBD observations from the current
step $t$ and node $i$, respectively. In our implementation,
$f_{\mathrm{PnP}}(\cdot)$ first extracts and matches features using
XFeat~\cite{potje2024xfeat} and LightGlue~\cite{lindenberger2023lightglue}
to obtain 2D-3D correspondences. We adopt this combination primarily for its efficiency, as both methods provide fast and reliable feature processing. The resulting correspondences are then passed to an E-PnP~\cite{lepetit2009ep} solver to estimate the relative pose.

Given this relative measurement, we obtain an estimate of the robot’s global pose by composing the keyframe pose $c_i$ with the relative transform, i.e., $x_t \approx c_i\, T_{i,t}$. 
The distribution of the composed pose is obtained by pushing forward each mixture component by $T_{i,t}$, analogous to the operation used in the motion factor. The resulting global pose distribution conditioned on keyframe $i$ is therefore a Gaussian mixture over the $K$ components of $c_i$:
\begin{equation}
p(x_t \mid v_i)
\approx
\sum_{k=1}^K
w_i^{(k)}\,
\mathcal{N}_{\mathfrak{se}(3)}
\!\left(
    \log\!\bigl( (\mu_i^{(k)} T_{i,t})^{-1} x_t \bigr);
    \; 0,\Sigma_{i,t}^{(k)}
\right),
\label{eq:composition-gmm}
\end{equation}
where
$
\Sigma^{(k)}_{i,t}
\approx
\mathrm{Ad}_{T_{i,t}^{-1}}\,
\Sigma^{(k)}_{i}\,
\mathrm{Ad}_{T_{i,t}^{-1}}^{\top}
+
\Gamma^{\text{pnp}}_{i,t}
$.
Here, $\Gamma^{\text{pnp}}_{i,t}$ is the relative-pose covariance from PnP in $\mathfrak{se}(3)$; we approximate it using the PnP inlier support via a fixed isotropic base covariance scaled by the inlier count,
$\Gamma^{\text{pnp}}_{i,t} = \lambda I_6 / \max(n^{\text{in}}_{i,t},1)$.
Eq.~\eqref{eq:composition-gmm} describes the pose likelihood conditioned on a specific
keyframe $i$. However, multiple keyframes may be visually compatible with the
current observation, making data association an inherent component of the
measurement update. To model this uncertainty, we introduce a discrete latent
variable $Y_t \in \{1,\dots,N_t\}$ indicating which keyframe explains $z_t$.
Marginalizing over this association yields a mixture-form measurement message:
$
\mathcal{H}_t(x_t)=\sum_{i=1}^{N_t}
p(x_t \mid z_t, Y_t{=}i, \mathcal{G})\,
p(Y_t{=}i \mid z_t, \mathcal{G}),
$
subject to $\sum_{i=1}^{N_t} p(Y_t{=}i \mid z_t,\mathcal{G}) = 1$.
The association term $p(Y_t{=}i \mid z_t,\mathcal{G})$ is estimated using a compact proposal-weighting approximation. 
For each retrieved keyframe, we combine two cues: the VPR score, which captures appearance-level compatibility, and the PnP inlier count, which captures whether a geometrically consistent relative pose can be estimated. 
In our system, place recognition is performed by BoQ~\cite{ali2024boq}, while relative pose estimation is obtained via RANSAC E-PnP~\cite{lepetit2009ep}. 
We compute an association weight
$
P(i)
=
\mathrm{softmax}\!\left(
    \pi(i)\,\frac{\mathrm{inlier}(i)}{M}
\right),
$
where $\pi(i)$ is the VPR retrieval score, $\mathrm{inlier}(i)$ denotes the PnP inlier count, and $M$ is the maximum number of detected features used for normalization. 
Empirically, this appearance--geometry weighting provides a reliable approximation to
$p(Y_t=i \mid z_t,\mathcal{G})$ for ranking candidate associations. 
The subsequent product with the motion message and delayed loop-closure test provide the main geometric and temporal validation of these proposals.
Using $P(i)$, the global measurement message takes the
mixture form:
\begin{equation}
\small
\mathcal{H}_t(x_t)
=
\sum_{i=1}^{N_t} P(i)
\sum_{k=1}^{K} w_i^{(k)}
\,
\mathcal{N}_{\mathfrak{se}(3)}
\!\left(
    \log\!\bigl(
        (\mu_i^{(k)} T_{i,t})^{-1} x_t
    \bigr);
    \; 0,\Sigma_{i,t}^{(k)}
\right).
\label{eq:measurement}
\end{equation}
In contrast to conventional GSFs, which apply per-component EKF-style measurement updates, this global measurement message $\mathcal{H}_t$ aggregates evidence over all retrieved keyframes and enables the filter to naturally handle tracking loss, and loop closures, since the measurement update is not restricted to refining only the currently dominant mixands.

\vspace{-0.5em}
\mypara{Hypothesis Management.} One challenge that arises from the above message passing is that the direct product produces up to $N_t K^2$ mixture terms. Such quadratic growth quickly becomes impractical in long sequences. Rather than computationally-expensive EM-style reduction~\cite{990138}, we adopt
an explicit {hypothesis management} strategy. 
Let $w^{(i)}_{\text{mot}}$ and $w^{(j)}_{\text{meas}}$ be the weights of the
$i$-th motion and $j$-th measurement components, respectively. The product
component $(i,j)$ has weight
$
w_t^{(i,j)}
\propto
w^{(i)}_{\text{mot}}
w^{(j)}_{\text{meas}}
C_{ij},
$
where $C_{ij}$ is their Gaussian overlap. Most pairs have negligible overlap because they are far apart in $\mathrm{SE}(3)$.
To control the growth in mixture size, we maintain a small set of physically meaningful {hypotheses}. Each surviving mixture component in $p(x_t \mid z_{1:t}, u_{1:t}, \mathcal{G})$ is interpreted as a hypothesis $h^{(l)}$ representing a distinct, dynamically consistent
trajectory. We summarize
$
h^{(l)}_t
=
\{
x^{(l)}_{1:t},
\Sigma^{(l)}_{1:t},
\bar w^{(l)}_t,
\mathcal{E}^{(l)}_{\text{vis}}
\},
$
where $x^{(l)}_{1:t}$ are the node poses along the trajectory,
$\Sigma^{(l)}_{1:t}$ are their covariances,
$\bar w^{(l)}_t$ is the normalized hypothesis weight, and
$\mathcal{E}^{(l)}_{\text{vis}}$ is the set of visual constraints (further described in the smoothing step in Section~\ref{sec:pgo-smoothing}). 

In brief, we manage hypotheses in two stages (complete details in Appendix~\ref{app:hyp_management}):
\begin{enumerate}[leftmargin=15pt,itemsep=1pt, topsep=0pt, parsep=0pt, partopsep=0pt]
\item \textbf{Measurement Clustering}, which clusters all measurement components on $\mathrm{SE}(3)$ into a small number of dominant modes using an $\mathrm{SE}(3)$-aware DBSCAN method; 
\item \textbf{Fusion, Birth, and Pruning}, whereby we fuse these modes with the motion mixture, updating existing hypotheses and spawning new ones if needed. In addition, we prune hypotheses whose weights or overlaps are too small.  
\end{enumerate}
In practice, we do not duplicate node poses; each hypothesis $h^{(l)}$ is represented solely by its component index $l$ at every node, and under the motion model this component evolves continuously in $\mathrm{SE}(3)$, so the index $l$ is preserved over time (the $l$-th component of $m_t^{\text{mot}}$ is the propagation of the $l$-th component at time $t{-}1$). New hypotheses are created only when the global measurement message proposes a mode that does not overlap with any existing motion component, at which point a new component index, and thus a new hypothesis, is branched out.

\begin{figure}[t]
    \centering

    \makebox[\linewidth][l]{\textbf{Rover (Campus)}}\\[1pt]
    \makebox[\linewidth][c]{%
        \includegraphics[width=0.162\linewidth]{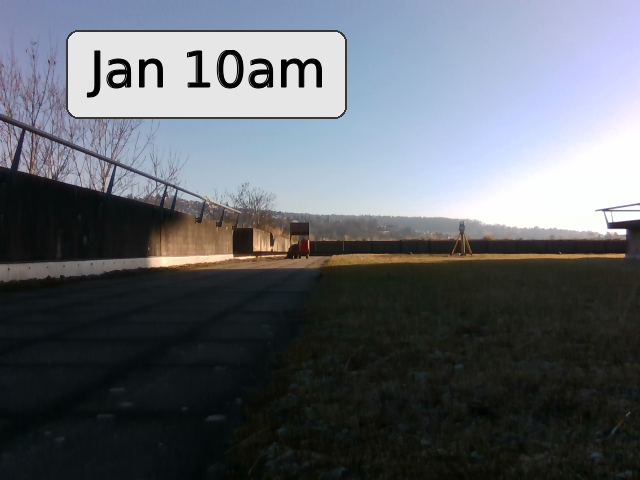}\hfill
        \includegraphics[width=0.162\linewidth]{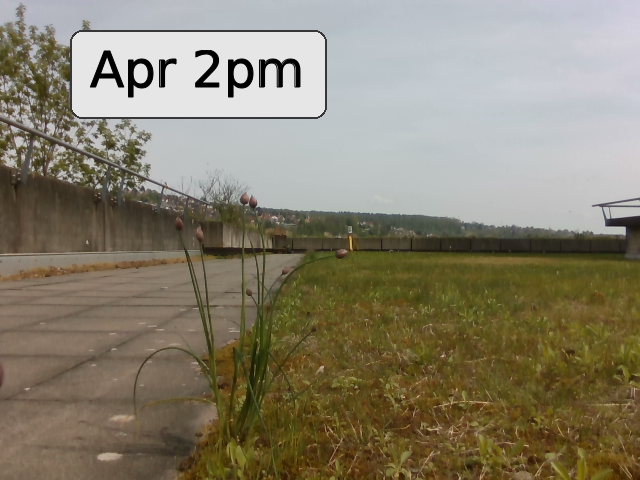}\hfill
        \includegraphics[width=0.162\linewidth]{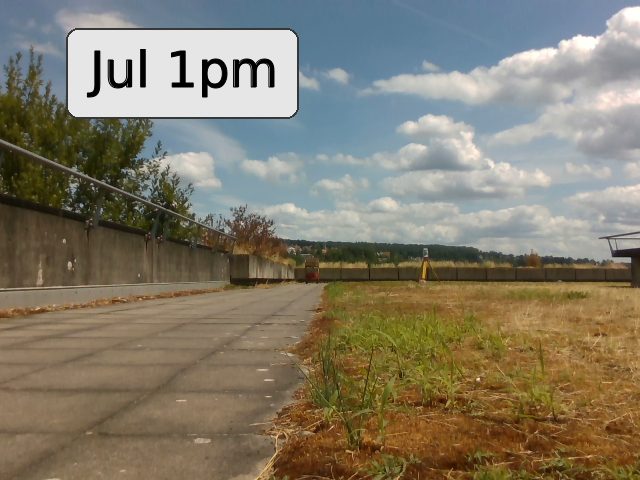}\hfill
        \includegraphics[width=0.162\linewidth]{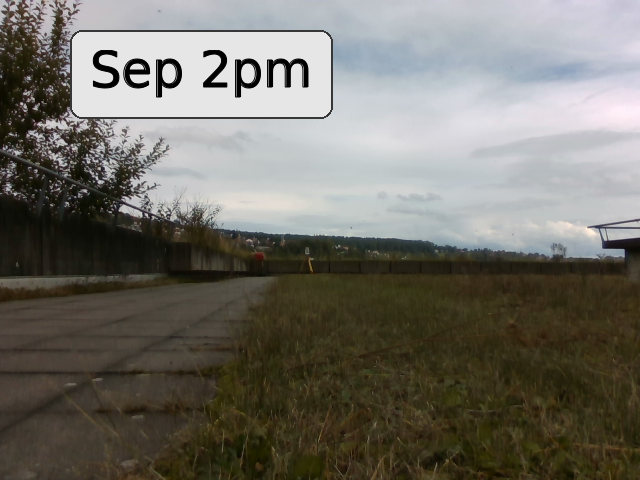}\hfill
        \includegraphics[width=0.162\linewidth]{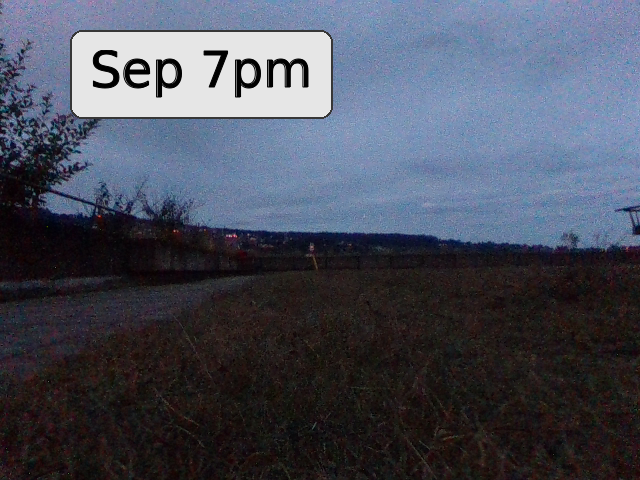}\hfill
        \includegraphics[width=0.162\linewidth]{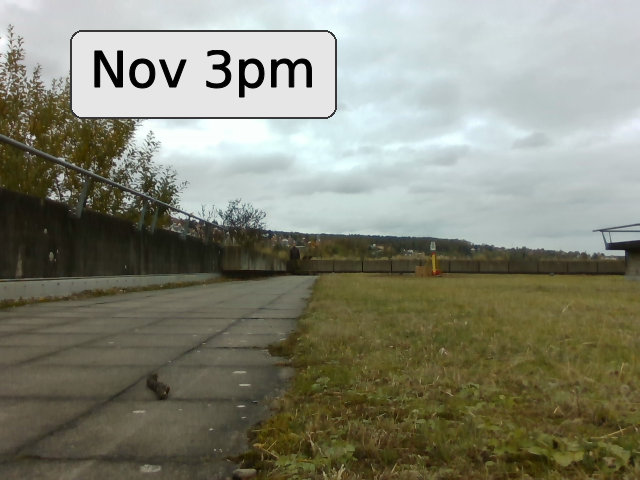}%
    }

    \vspace{4pt}

    \makebox[\linewidth][l]{\textbf{OpenLORIS (Corridor)}}\\[1pt]
    \makebox[\linewidth][c]{%
        \includegraphics[width=0.246\linewidth]{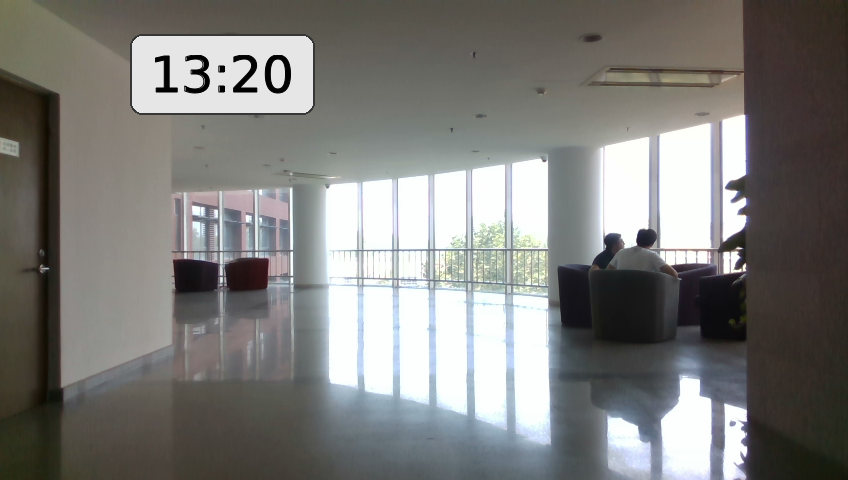}\hfill
        \includegraphics[width=0.246\linewidth]{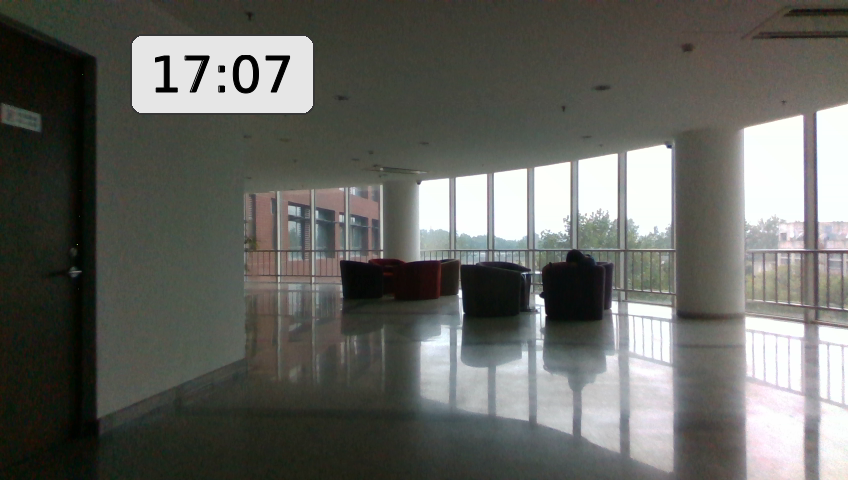}\hfill
        \includegraphics[width=0.246\linewidth]{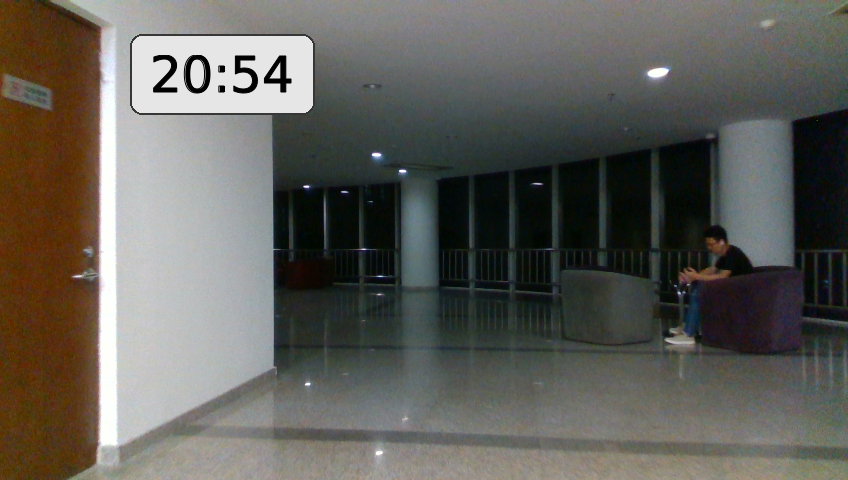}\hfill
        \includegraphics[width=0.246\linewidth]{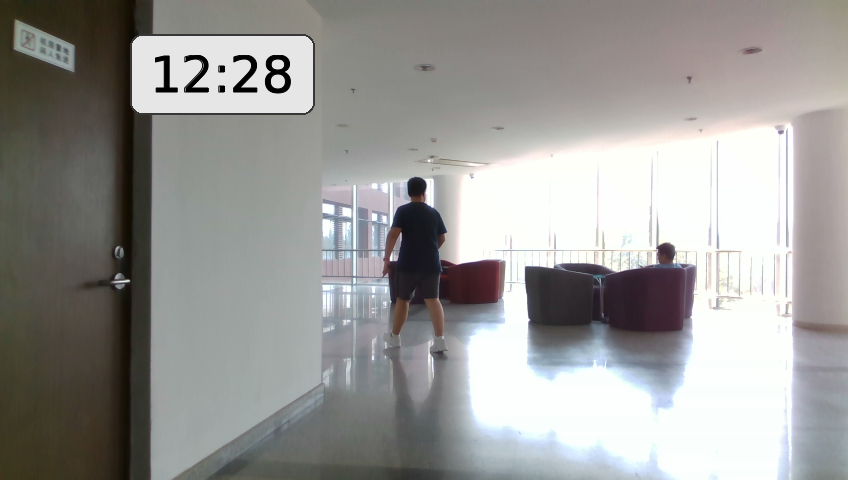}%
    }

    \caption{\small Appearance change at the same physical locations for the Rover and OpenLORIS  benchmarks. 
    }
    \label{fig:appearance_change}
\end{figure}

\vspace{-0.5em}
\subsection{Topological Map Management}
\label{sec:map-management}
Given the state estimates, we maintain a topological map. This involves creating new nodes, adding edges, and performing pose-graph optimization (PGO) to smooth past trajectories.

\vspace{-0.5em}
\mypara{Node Creation.}
At each timestep, we first perform visual place recognition (VPR) to retrieve keyframe candidates whose appearances match the current observation. We create a new node whenever the maximum similarity score falls below a threshold $\beta$, using the current RGB-D frame and its pose estimate (e.g., the robot visits a previously unseen location or when a known location exhibits a substantially different appearance).
Since the current pose estimate is represented as a Gaussian mixture on $\mathrm{SE}(3)$, the newly added node inherits the same mixture structure, i.e., a set of mixture components $\{(\mu_{\text{new}}^{(k)}, \Sigma_{\text{new}}^{(k)}, w_{\text{new}}^{(k)})\}_{k=1}^{K}$,
where each component is parameterized by a pose mean
$\mu_{\text{new}}^{(k)} \in \mathrm{SE}(3)$ and diagonal covariance $\Sigma_{\text{new}}^{(k)}$ defined in the tangent space $\mathfrak{se}(3)$.

\vspace{-0.5em}
\mypara{Edge Creation.}
Whenever a new keyframe is created, we add two types of edges:
\begin{itemize}[leftmargin=15pt,noitemsep,topsep=0pt]
    \item \textbf{Odometry edge:} connecting the new node to the previous node, induced by the physical motion between timesteps.
    \item \textbf{Proximity edge:} connecting the new node to nearby existing nodes within a spatial radius $d$.  This helps ensure that multiple nodes representing the same physical place (e.g., under different appearances) remain connected in the topological graph.  
    In principle, a learned traversability model~\cite{wang2025genie} could be used to validate 
    whether two nodes should be connected, but in practice we find that a simple distance threshold (e.g., $d = 0.5\,\mathrm{m}$) is reliable and lightweight.
\end{itemize}

\phantomsection
\label{sec:pgo-smoothing}

\vspace{-0.5em}
\mypara{Smoothing.}
While our representation is topological, we periodically refine node poses to support accurate temporal hypothesis filtering. This refinement is computationally inexpensive because the map contains no explicit point landmarks. We estimate node poses by optimizing the posterior, 
$
p\!\left(\{c_i\}\mid
\mathcal{E}_{\text{odo}},
\mathcal{E}_{\text{vis}},
z_{1:t},u_{1:t}\right),
$
where $\mathcal{E}_{\text{odo}}$ is the set of odometry edges and
$\mathcal{E}_{\text{vis}}$ is the set of visual constraints. Each edge encodes a noisy relative pose in
$\mathrm{SE}(3)$; under a Gaussian noise model, the negative log-posterior is a
sum of squared residuals in $\mathfrak{se}(3)$. 

For each hypothesis $h^{(l)}$ we maintain $\mathcal{E}^{(l)}_{\text{vis}}
=
\{\, (i,j) \,\},
$
where $(i,j)$ denotes a visual constraint between node $i$ and node $j$ for hypothesis $l$. Under our hypothesis management policy, all visual constraints in $\mathcal{E}^{(l)}_{\text{vis}}$ implicitly connect component $l$ at both endpoints. Conditioning on $h^{(l)}$ therefore fixes the component identities, and the continuous part of the problem reduces to a standard pose-graph optimization (PGO) over the node means. Thus, we minimize the negative log posterior 
$p(\{c_i^{(l)}\}\mid \mathcal{E}_{\text{odo}},\mathcal{E}^{(l)}_{\text{vis}},
z_{1:t},u_{1:t})$.  
Hypotheses whose optimized pose graphs are inconsistent with odometry or other constraints naturally receive
lower posterior weight and are subsequently pruned. 

\vspace{-1em}
\mypara{Loop Closure.}
Loop closure detection in our system arises naturally from the temporal hypothesis filtering (THF) formulation: a revisit, whether a standard loop
closure or a kidnapped-robot event, appears as an additional hypothesis whose
trajectory becomes consistent with an older region of the map. After the THF step, we retain a small set of hypotheses $\{h_t^{(l)}\}_{l=0}^{L-1}$ with normalized weights $\{\bar w_t^{(l)}\}_{l=0}^{L-1}$, where $h_t^{(0)}$ is the
currently tracked branch. For each alternative hypothesis $l>0$, we compute a
log posterior odds against the null hypothesis $h^{(0)}$:
$\ell_t^{(l)} = \log \bar w_t^{(l)}/\bar w_t^{(0)}$.
A positive $\ell_t^{(l)}$ indicates that, at time $t$, hypothesis $h^{(l)}$
is more likely than the no-loop hypothesis $h^{(0)}$. Rather than accepting a
loop closure from a single frame, we accumulate evidence over a sliding window
of length $W$ by counting how many times $\ell_\tau^{(l)} > 0$ for
$\tau \in \{t-W+1,\dots,t\}$. A loop closure for hypothesis $h^{(l)}$ is
declared only when this count exceeds a threshold $r$. This suppresses
spurious place-recognition outliers and ensures that only hypotheses with
sustained support are promoted.

Once a loop closure between $h^{(0)}$ and $h^{(l)}$ is accepted, we perform a joint PGO analogous to the smoothing procedure, but over both hypotheses simultaneously. For each node index $i$ in the temporal overlap of the two trajectories, we introduce a loop-closure residual
$\log((\mu_i^{(0)})^{-1} \mu_i^{(l)})$,
with a small covariance, enforcing consistency between the two pose estimates
that correspond to the same physical location.
We then construct an augmented PGO whose variables are the union of node poses
$\{\mu_i^{(0)}\} \cup \{\mu_i^{(l)}\}$, and whose factors include:
(i) the odometry and visual constraints from each hypothesis, and
(ii) the identity loop-closure constraints
$\mathcal{E}^{(0,l)}_{\text{lc}}$ linking corresponding nodes.
Solving this joint optimization aligns the two branches and yields a single,
globally consistent trajectory. 

After optimization, we collapse the merged pair into a single hypothesis by
retaining index $0$ with updated poses and aggregated weight
$\bar w_t^{(0)} \leftarrow \bar w_t^{(0)} + \bar w_t^{(l)}$, and removing
$h^{(l)}$ from the mixture. In this way, loop closures and  tracking losses are realized as
hypothesis mergers driven by the same THF machinery, yielding a unified and robust treatment of
revisits under strong perceptual aliasing.

\newcommand{\heatimg}[2]{\includegraphics[width=0.49\linewidth]{figures/exp/heatmap_plot/#1/#2.png}}
\newcommand{\barimg}[2]{\includegraphics[width=\linewidth]{figures/exp/barcode_plot/#1/#2.png}}

\begin{figure}[t]
    \centering
    \includegraphics[width=0.99\linewidth]{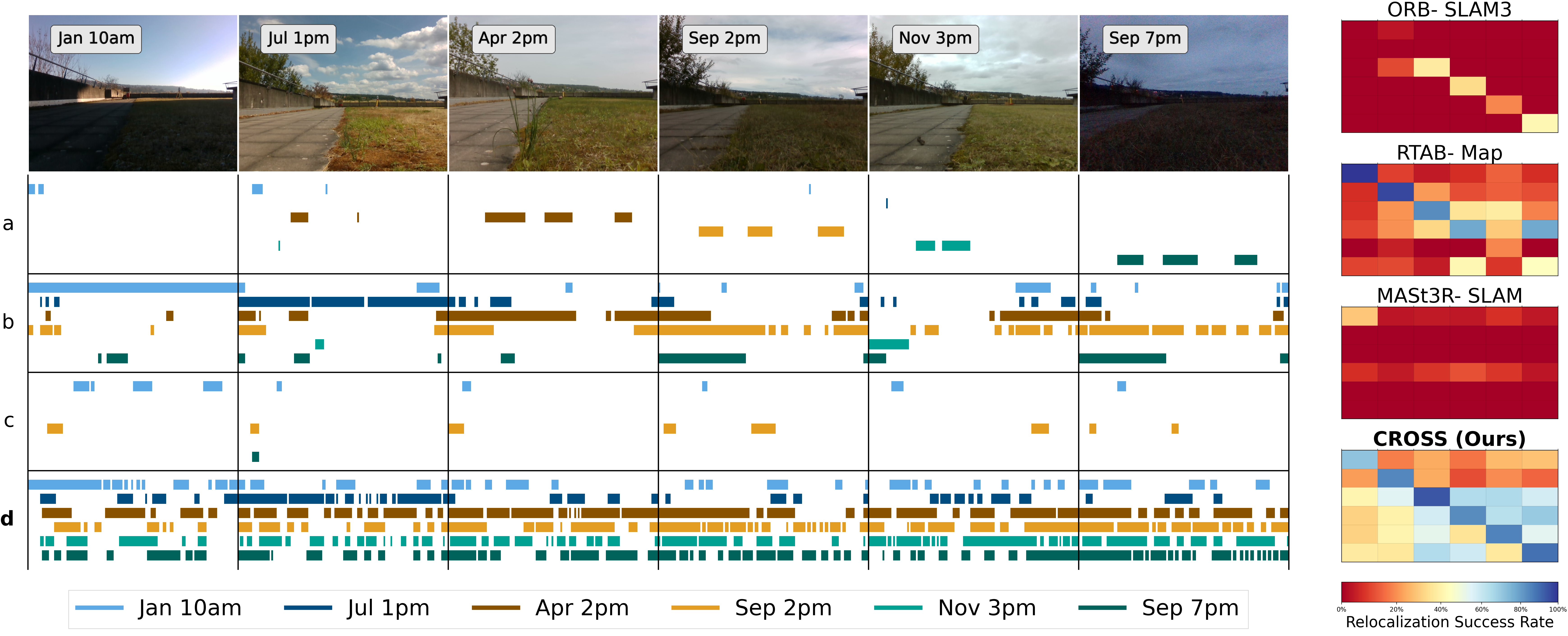}
\caption{\small{Multi-session relocalization results on the Rover~\cite{schmidt2025rover} Campus scene.
\textbf{Left:} Relocalization outcomes across different locations.
Each row corresponds to a mapping trajectory (indicated by different colors), while columns
show relocalization attempts at the same physical locations captured at different
times or months, as illustrated in the top image.
Empty space indicates relocalization failed at that specific location.
The compared methods are:
(a) ORB-SLAM3,
(b) RTAB-Map,
(c) MASt3R-SLAM,
and (d) Ours.
Most baseline methods struggle under significant lighting and appearance changes,
whereas our approach consistently relocalizes despite substantial visual variation.
\textbf{Right:} A compact summary, where each grid cell reports the
relocalization success rate for a given mapping--testing sequence pair.
Full results and additional analysis are provided in
Section~\ref{subsec:exp-appearance-change}.}}
    \label{fig:campus_large}
    \vspace{-1em}
\end{figure}

\section{Experimental Evaluation}
\label{sec:experiments}
\vspace{-0.5em}
The objective of our experiments is to test whether CROSS can localize and act using a previously built spatial memory despite realistic environmental change. We evaluate this under three representative condition shifts: (i) illumination changes across time of day, (ii) long-term appearance changes across months, and (iii) object-level changes caused by rearrangement or motion.

Additional experiments and uncertainty analyses in Appendix~\ref{app:further_exps} evaluate perceptual aliasing, fast motion, occlusion, noisy odometry, runtime, and Wilson confidence intervals for the binary success-rate metrics, showing that CROSS remains stable under degraded sensing and runs in real time.

\subsection{Appearance Change Across Time}
\label{subsec:exp-appearance-change}

\vspace{-1em}
\mypara{Datasets.}
We evaluate \emph{change-robustness} under appearance variation using two public datasets (Figure~\ref{fig:appearance_change}). The OpenLORIS~\cite{shi2019openlorisscene} \emph{Corridor} scene is an indoor dataset with repeated traversals captured at different times of day, exhibiting substantial illumination variation along an approximately 80\,m corridor.
The Rover~\cite{schmidt2025rover} \emph{Campus} scene is an outdoor dataset comprising repeated traversals collected across different months and times of day, covering an approximately 270\,m route and capturing long-term appearance changes due to lighting, weather, and seasonal effects.

\vspace{-0.5em}
\mypara{Methods.}
We compare our approach against representative and state-of-the-art SLAM systems, including ORB-SLAM3 (RGBD)~\cite{campos2021orb}, RTAB-Map~\cite{labbe2019rtab}, and MASt3R-SLAM~\cite{murai2025mast3r}. In addition, we evaluate common topological localization baselines: Greedy Matching (GM), Sequence Matching (SM), Probabilistic Belief Update (PBU), and Appearance-based Mapping~\cite{labbe2013appearance}. GM, SM, and PBU represent widely used strategies for online topological mapping and localization~\cite{savinov2018semi,meng2020scaling,suomela2024placenav}. Several methods are not publicly available, so we reimplemented the topological baselines using the same VPR models and comparable hyperparameters to ensure a fair comparison (see Appendix~\ref{app:topo-baselines}).

\begin{wraptable}[14]{r}{0.5\linewidth}
\vspace{-0.5cm}
\centering
\caption{\footnotesize Relocalization success rate (RS) under appearance change on indoor (OpenLORIS) and outdoor (Rover) benchmarks, without vs.\ with self mapping.}
\label{tab:appearance-change}

\scriptsize
\setlength{\tabcolsep}{2.4pt}
\renewcommand{\arraystretch}{0.88}

\begin{adjustbox}{max width=\linewidth}
\begin{tabular}{@{}lcccccc@{}}
    \toprule
    & \multicolumn{3}{c}{\textbf{cross-sequence}} 
    & \multicolumn{3}{c}{\textbf{w.\ self mapping}} \\
    \cmidrule(lr){2-4} \cmidrule(lr){5-7}
    \textbf{Method} &
    \textbf{\makecell{RS$\uparrow$\\(OLS)}} &
    \textbf{\makecell{RS$\uparrow$\\(Rover)}} &
    \textbf{\makecell{RS$\uparrow$\\(All)}} &
    \textbf{\makecell{RS$\uparrow$\\(OLS)}} &
    \textbf{\makecell{RS$\uparrow$\\(Rover)}} &
    \textbf{\makecell{RS$\uparrow$\\(All)}} \\
    \midrule

    \multicolumn{7}{@{}l}{SLAM Systems} \\
    ORB3~\cite{campos2021orb}      & 0.032 & 0.005 & 0.013 & 0.003 & 0.048 & 0.040 \\
    RTAB~\cite{labbe2019rtab}      & 0.292 & 0.158 & 0.198 & 0.438 & 0.275 & 0.336 \\
    MASt3R~\cite{murai2025mast3r}  & 0.173 & 0.014 & 0.062 & 0.226 & 0.024 & 0.094 \\
    \midrule

    \multicolumn{7}{@{}l}{Topological Methods} \\
    GM                             & 0.080 & 0.064 & 0.069 & 0.483 & 0.263 & 0.345 \\
    SM                             & 0.073 & 0.152 & 0.128 & 0.479 & 0.333 & 0.387 \\
    PBU                            & 0.080 & 0.063 & 0.068 & 0.482 & 0.262 & 0.344 \\
    ABM~\cite{labbe2013appearance} & 0.231 & 0.020 & 0.090 & \textbf{0.569} & 0.229 & 0.356 \\
    \textbf{CROSS}~(Ours)          & \textbf{0.353} & \textbf{0.397} & \textbf{0.384} & 0.478 & \textbf{0.494} & \textbf{0.488} \\
    \bottomrule
\end{tabular}
\end{adjustbox}
\end{wraptable}

\vspace{-0.5em}
\mypara{Setup and Metrics.}
We evaluate all methods under a map--query protocol. A single traversal is first designated as the \emph{mapping sequence}, during which the system incrementally constructs its spatial representation online. Relocalization performance is then evaluated on a separate traversal recorded under different environmental conditions (the \emph{testing sequence}). The testing sequence is partitioned into fixed-length subsequences of 200 frames, each treated as an online relocalization trial. This protocol results in 735 trials and 1,232 trials for OpenLORIS and Rover, respectively.

We report the \textit{relocalization success} (RS) rate as the primary evaluation metric. For methods that output a discrete topological estimate (e.g., node or place ID), a trial is considered successful if the predicted location lies within a distance threshold $r_D$ of the ground-truth position. For metric SLAM systems that output a full pose estimate, success is defined as the estimated pose lying within the same threshold $r_D$ of the ground-truth pose. We set $r_D=2\,\mathrm{m}$ for indoor environments and $r_D=5\,\mathrm{m}$ for outdoor environments. We use two evaluation settings: \emph{with self-mapping}, where the mapping and testing sequences are identical, and \emph{cross-sequence}, where relocalization is performed using maps built from different sequences. 

\vspace{-0.5em}
\mypara{Results.}
Overall, CROSS achieves the highest relocalization success rates across both indoor and outdoor benchmarks (Table~\ref{tab:appearance-change}), indicating improved robustness to appearance change compared to existing SLAM-based and topological methods. Wilson confidence intervals for all success rates are provided in Appendix~\ref{app:statistical-uncertainty}. Absolute performance remains limited under severe condition shifts: even for CROSS, success rates in cross-sequence settings are around 40\%. Performance degrades substantially under large appearance gaps, such as day-to-night transitions or long-term seasonal changes (e.g., January to September). While non-perfect success rates are acceptable in practice due to repeated relocalization opportunities as the robot moves, higher success rates directly translate to greater tolerance to appearance change and faster recovery. 

\subsection{Object Navigation on a Real Quadruped Robot}
\label{subsec:real-robot}

To evaluate whether the spatial memory facilitates downstream semantic navigation, we deployed CROSS on a quadruped robot in a changing indoor environment. 
This experiment is not intended as a benchmark of open-vocabulary semantic reasoning; instead, it tests whether our stored visual memory can be reused for localization and object-goal navigation under conditions commonly encountered in real-world deployment.

\vspace{-0.5em}
\mypara{Task and Setup.}
During an initial \emph{mapping session}, the robot constructs the spatial memory. In a subsequent \emph{query session}, the robot is initialized at a random location within the same environment and tasked with \emph{navigating to a specified target object} (e.g., a bin, plant, or coke bottle) that was previously observed.

\begin{wrapfigure}[13]{r}{0.5\columnwidth}
\vspace{-0.3cm}
    \centering
    \includegraphics[width=0.32\linewidth]{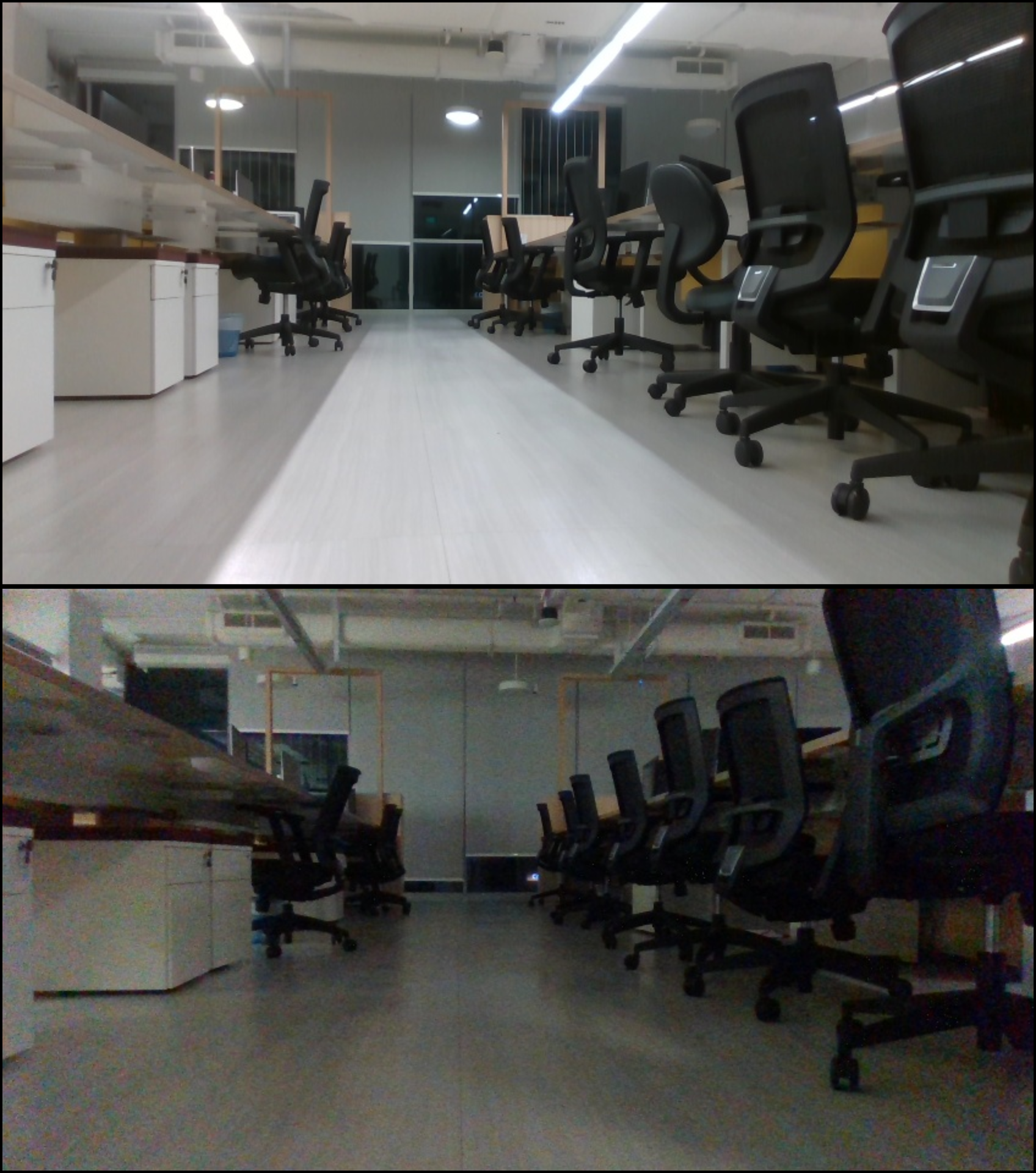}\hfill
    \includegraphics[width=0.32\linewidth]{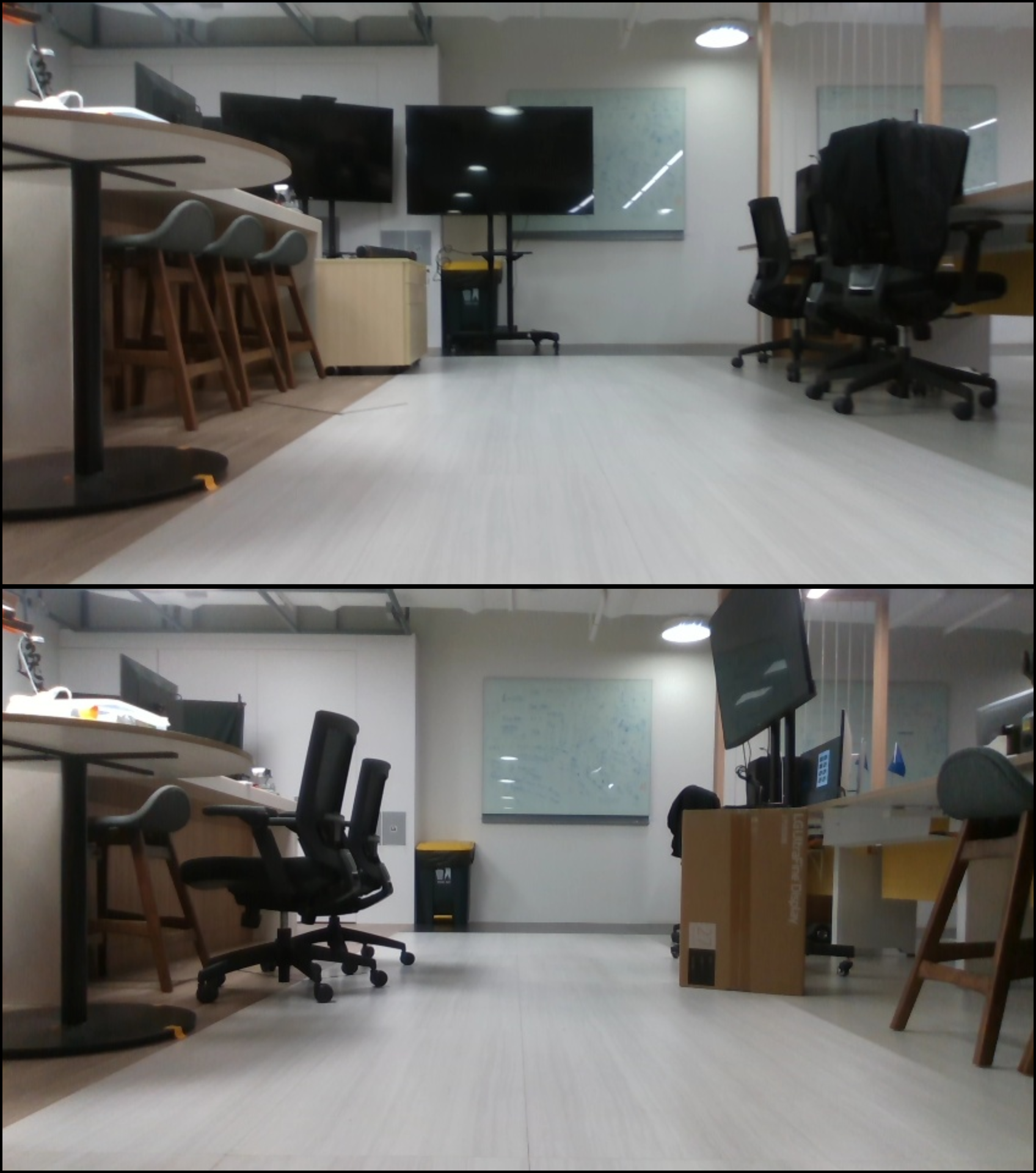}\hfill
    \includegraphics[width=0.32\linewidth]{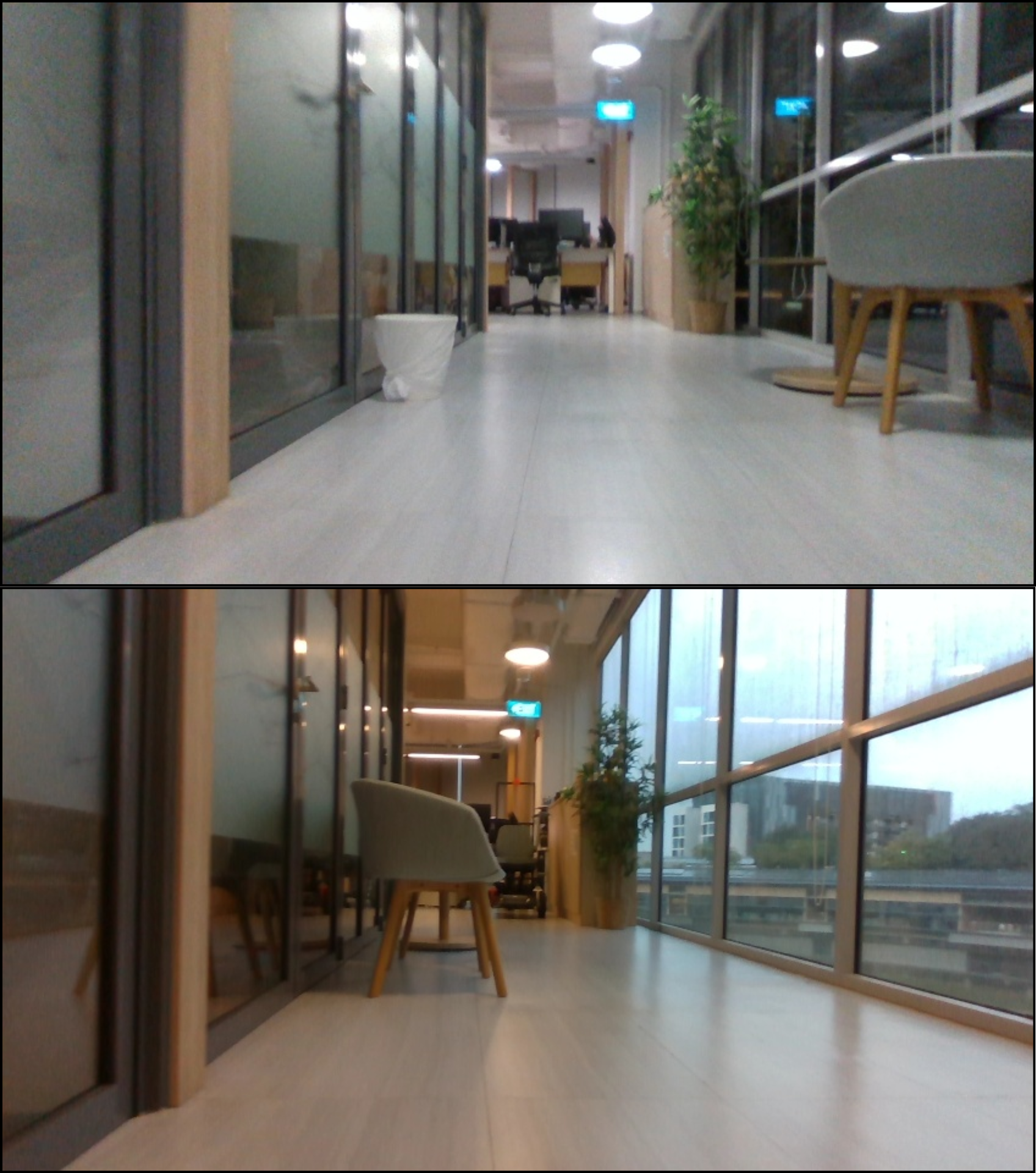}

    \caption{\small
    Example evaluation settings from the real quadruped-robot experiment, with each scene shown \emph{before} (top) and \emph{after} (bottom) the change.
    \textbf{Left:} Lighting Change (LC).
    \textbf{Middle:} Object Rearrangement (OR).
    \textbf{Right:} Combined Change (LC+OR).
    }
    \label{fig:real_ssi_settings}
\end{wrapfigure}
We evaluate three experimental settings that capture common sources of environmental change. 
\emph{Lighting Change (LC)} considers only appearance variation induced by different times of day between the mapping and query sessions. 
\emph{Object Rearrangement (OR)} evaluates robustness to object-level changes, where furniture and objects are moved while lighting conditions remain similar. 
\emph{Combined Change (LC+OR)} includes both time-of-day variation and object rearrangement simultaneously. Figure~\ref{fig:real_ssi_settings} illustrates representative examples of the LC, OR, and LC+OR settings.
For each setting, we conduct 10 independent trials with different scene configurations and target objects. 
A trial is considered successful if the robot reaches the specified target object within a distance of 1\,m.
\vspace{-0.5em}

\mypara{Methods.}
We compare CROSS against the commonly used metric map + semantic layer \textbf{(M+S)} paradigm, which augments a metric SLAM map with an additional semantic layer, as adopted in prior work such as~\cite{chaplot2020object, chang2024goat, gu2024conceptgraphs}. We implement two variants of this baseline, using ORB-SLAM3~\cite{campos2021orb} and RTAB-Map~\cite{labbe2019rtab} for localization and mapping. Implementation details are provided in Appendix~\ref{app:real-robot-implementation}.

\begin{wraptable}[8]{r}{0.40\linewidth}
\vspace{-0.5cm}
\centering
\caption{\small Success rate (\%) in real-robot object-goal navigation under environmental changes.}
\label{tab:real_robot_results}
\footnotesize
\setlength{\tabcolsep}{3pt}
\begin{tabularx}{\linewidth}{@{}l c c c@{}}
\toprule
Method & LC & OR & LC+OR \\
\midrule
M+S (ORB-SLAM3) & 30.0 & 30.0 & 20.0 \\
M+S (RTAB-Map) & 40.0 & 60.0 & 30.0 \\
CROSS (Ours) & \textbf{70.0} & \textbf{80.0} & \textbf{80.0} \\
\bottomrule
\end{tabularx}
\end{wraptable}

\vspace{-0.5em}
\mypara{Results.}
Across all three settings, CROSS consistently achieves a higher task success rate than the baseline (Table~\ref{tab:real_robot_results}), which demonstrates that the proposed representation can effectively support object-goal navigation under both appearance and object-level changes. 

We conducted a qualitative real-robot experiment to examine if CROSS remains reliable in the presence of dynamic agents (i.e., moving pedestrians) across different times of the day (Fig.~\ref{fig:teaser}). We performed a mapping session over a $\approx$170\,m trajectory covering walkways and a university canteen.
We then ran a query session when the same area was crowded with different lighting conditions, and commanded the robot to navigate to a language-specified target (security post).
Despite substantial scene changes, the robot re-localized and completed the navigation task. 


\section{Conclusion and Future Work}
\label{sec:conclusion}
We presented CROSS, a change-robust online topological spatial memory for long-term semantic navigation. 
CROSS stores raw RGB-D keyframes as task-agnostic visual observations and maintains a bounded Gaussian-mixture belief over continuous $\mathrm{SE}(3)$ poses. 
By constructing global visual observation modes and validating them through motion consistency over time, CROSS can recover from tracking loss, reject many perceptual aliases, and promote consistent branches to loop-closure constraints. 
Experiments on public indoor/outdoor benchmarks and real-robot object-goal navigation show improved reuse of a single memory under appearance and object-level change.

\mypara{Limitations and Future Work.}
CROSS relies on reliable relative-pose estimates between the current RGB-D frame and retrieved keyframes. 
Our current implementation uses keypoint matching and PnP, which can fail under extreme appearance shift, low overlap, motion blur, or noisy depth; the filtering formulation itself can also use learned relative-pose estimators when they become sufficiently accurate and efficient. 
A second limitation is semantic scope: this work uses stored observations and object-query annotations/detections to demonstrate semantic navigation, rather than evaluating full open-vocabulary spatial reasoning. 
Future work will improve relative-pose estimation under large visual change and evaluate richer language-conditioned spatial queries.

\clearpage
{\small
\bibliographystyle{plainnat}
\bibliography{references}
}

\clearpage
\appendix

\section{Experiment Details}
\subsection{Topological Localization Baselines}
\label{app:topo-baselines}

This appendix describes the topological localization baselines used in our experiments. All methods operate on the same topological graph and use the same underlying VPR model~\cite{ali2024boq} for computing visual similarity scores, differing only in how temporal information and motion constraints are incorporated.

\subsubsection{Greedy Matching (GM)}

The greedy matching baseline localizes by selecting the node with the highest similarity score to the current observation. If the maximum similarity exceeds a fixed threshold $\tau$, the corresponding node is selected as the localization result; otherwise, the localization estimate remains unchanged. This baseline reflects a common retrieval-based relocalization strategy without temporal reasoning.

\subsubsection{Sequence Matching (SM)}

Instead of matching a single observation, sequence matching aggregates similarity scores over a short temporal window to improve robustness against perceptual aliasing and viewpoint variation. A candidate match between nodes $(v_i, v_j)$ is accepted if the aggregated similarity over a window of size $2h{+}1$ satisfies
\[
\mathrm{f}\!\left(
\mathrm{sim}(z_{v_i-h}, z_{v_j-h}), \ldots,
\mathrm{sim}(z_{v_i+h}, z_{v_j+h})
\right) \ge \tau,
\]
where $\mathrm{sim}(\cdot,\cdot)$ denotes the visual similarity score and $\mathrm{f}(\cdot)$ is an aggregation function. In our implementation, we use the median of the similarity scores within the window, which provides robustness to outliers. This baseline is inspired by prior work on sequence-based place recognition~\cite{savinov2018semi,meng2020scaling}.

\subsubsection{Probabilistic Belief Update (PBU)}

The probabilistic belief update baseline maintains a discrete posterior belief $b_t(v) = P(v_t = v \mid z_{1:t})$ over the topological nodes $v \in \mathcal{V}$ at time $t$. Given the belief at the previous timestep, the state is first propagated via a motion model $P(v_t \mid v_{t-1})$ that constrains allowable transitions based on the graph topology:
\[
P(v_t \mid v_{t-1}) \propto
\begin{cases}
\alpha & \text{if } \mathrm{dist}_{G}(v_t, v_{t-1}) \le w_u, \\
\beta  & \text{otherwise},
\end{cases}
\]
where $\mathrm{dist}_{G}(\cdot,\cdot)$ denotes the hop distance in the graph, $w_u$ is the maximum allowed movement per timestep, and $\alpha, \beta$ are constants controlling transition likelihoods. 

The predicted belief is then updated using the retrieval similarity scores as the observation likelihood $P(z_t \mid v_t) \propto g(\mathrm{sim}(z_t, z_{v_t}))$, where $g(\cdot)$ maps similarity scores to likelihood values. The recursive update is formulated as:
\[
b_t(v_t) = \eta \cdot P(z_t \mid v_t) \sum_{v_{t-1} \in \mathcal{V}} P(v_t \mid v_{t-1}) b_{t-1}(v_{t-1}),
\]
where $\eta$ is a normalization constant. This discrete Bayesian filtering suppresses spurious matches that violate motion constraints and improves robustness to perceptual aliasing, following prior topological localization approaches~\cite{suomela2024placenav}.

\subsection{Real robot experiment implementation}
\label{app:real-robot-implementation}

We design the real robot experiments to test relocalization and object goal navigation, under controlled combinations of appearance change and object rearrangement.
To prevent differences in environment coverage, the spatial-semantic representation for all methods is constructed from data collected along an identical mapping trajectory in each scene.
The navigation policy performs local obstacle avoidance and relative waypoint tracking, and does not consider the downstream impact of navigation actions on the state estimation.
This setup allows us to attribute success or failure rates to the robustness of the underlying representation, rather than to planning or navigation strategies.
We augment the default map representation of ORB-SLAM3~\cite{campos2021orb} to include object memory by associating the recorded RGB observations to computed map poses via timestamp alignment.
RTAB-MAP in its default implementation provides the associated RGB image for each map pose.
Consequently, all methods considered for evaluation support semantic queries, which we use for the object navigation task.
We retain the default tunable parameter values for all methods, to maintain fairness in comparison and avoid method specific tuning.


\begin{table}[t]
\centering
\caption{Localization performance under perceptual aliasing on \emph{Topo-Bench}. Results are reported for: \emph{Ambiguous + Positive} (A+P), \emph{Positive Only} (P.O.), \emph{Ambiguous Only} (A.O.) and \emph{Balanced Localization Accuracy} (BLA), defined as the geometric mean over the three regimes.}
\label{tab:topo_bench}
\begin{tabular}{lcccc}
\toprule
Method & A+P & P.O. & A.O. & BLA \\
\midrule
RTAB-Map~\cite{labbe2019rtab}         & 0.059 & \textbf{0.433} & \textbf{1}     & 0.301  \\
ORB-SLAM3~\cite{campos2021orb}        & 0.059 & 0.183 & \textbf{1}     & 0.227 \\
ABM~\cite{labbe2013appearance} & 0.02  & 0.328 & \textbf{1}     & 0.201  \\
OpenFabmap       & 0     & 0.112 & 0.732 & 0.0748 \\
Rat-Slam         & 0.02  & 0.097 & 0.443 & 0.103 \\
GM               & 0.078 & 0.302 & 0.959 & 0.288 \\
SM               & 0.118 & 0.187 & 0.959 & 0.281 \\
PBU              & 0.078 & 0.302 & 0.959 & 0.288 \\
\textbf{CROSS (Ours)}             & \textbf{0.275} & 0.336 & 0.99  & \textbf{0.452} \\
\bottomrule
\end{tabular}
\end{table}

\section{Additional Experiments, Results, and Analysis}
\label{app:further_exps}

\subsection{Statistical Uncertainty}
\label{app:statistical-uncertainty}

All main success metrics are binary trial outcomes. A relocalization trial either succeeds under the distance threshold defined in Section~\ref{subsec:exp-appearance-change}, and a real-robot object-navigation trial either reaches the target object within $1\,\mathrm{m}$ as defined in Section~\ref{subsec:real-robot}. We therefore report two-sided $95\%$ Wilson score confidence intervals for Bernoulli proportions. For a success rate $\hat p=k/n$, the interval is
\[
\frac{\hat p+\frac{z^2}{2n}}{1+\frac{z^2}{n}}
\;\pm\;
\frac{z}{1+\frac{z^2}{n}}
\sqrt{
\frac{\hat p(1-\hat p)}{n}
+
\frac{z^2}{4n^2}
},
\qquad z=1.96 .
\]
This interval does not assume normally distributed errors and remains well-defined for small success counts. For the public relocalization benchmarks, we use the trial counts in Section~\ref{subsec:exp-appearance-change}: $n=735$ for OpenLORIS and $n=1232$ for Rover. Since Table~\ref{tab:appearance-change} reports rates rounded to three decimals, the intervals below are computed by rounding each reported proportion to the nearest integer count. For the real-robot study, each setting contains $n=10$ independent trials.

\begin{table}[h]
\centering
\caption{Approximate $95\%$ Wilson confidence intervals for relocalization success rates in Table~\ref{tab:appearance-change}. Values are percentages. OLS denotes OpenLORIS.}
\label{tab:appearance-ci}
\scriptsize
\setlength{\tabcolsep}{3pt}
\renewcommand{\arraystretch}{0.95}
\begin{tabular}{lcccc}
\toprule
Method &
CROSS OLS &
CROSS Rover &
Self OLS &
Self Rover \\
\midrule
ORB-SLAM3 &
$3.2\,[2.2,4.7]$ &
$0.5\,[0.2,1.1]$ &
$0.3\,[0.1,1.0]$ &
$4.8\,[3.7,6.1]$ \\
RTAB-Map &
$29.2\,[26.1,32.6]$ &
$15.8\,[13.9,18.0]$ &
$43.8\,[40.2,47.4]$ &
$27.5\,[25.1,30.1]$ \\
MASt3R-SLAM &
$17.3\,[14.8,20.2]$ &
$1.4\,[0.9,2.2]$ &
$22.6\,[19.7,25.7]$ &
$2.4\,[1.7,3.5]$ \\
GM &
$8.1\,[6.3,10.3]$ &
$6.4\,[5.2,7.9]$ &
$48.3\,[44.7,51.9]$ &
$26.3\,[23.9,28.8]$ \\
SM &
$7.2\,[5.6,9.3]$ &
$15.2\,[13.3,17.3]$ &
$47.9\,[44.4,51.5]$ &
$33.3\,[30.8,36.0]$ \\
PBU &
$8.1\,[6.3,10.3]$ &
$6.3\,[5.1,7.8]$ &
$48.2\,[44.6,51.8]$ &
$26.2\,[23.8,28.7]$ \\
ABM &
$23.1\,[20.2,26.3]$ &
$2.0\,[1.4,3.0]$ &
$56.9\,[53.3,60.4]$ &
$22.9\,[20.6,25.3]$ \\
\textbf{CROSS} &
$\mathbf{35.3\,[32.0,38.8]}$ &
$\mathbf{39.7\,[37.0,42.5]}$ &
$47.8\,[44.2,51.4]$ &
$\mathbf{49.4\,[46.6,52.2]}$ \\
\bottomrule
\end{tabular}
\end{table}

The intervals support the main conclusion that CROSS improves cross-sequence relocalization under appearance change. The separation is strongest on Rover, where the interval for CROSS is well separated from all baselines. On OpenLORIS cross-sequence relocalization, CROSS has the highest point estimate, but its interval overlaps with RTAB-Map; we therefore interpret this result as improved average performance rather than definitive per-trial dominance. In the self-mapping setting, ABM has the highest OpenLORIS point estimate, while CROSS has the highest Rover point estimate and the highest aggregate rate in Table~\ref{tab:appearance-change}.

\begin{table}[h]
\centering
\caption{$95\%$ Wilson confidence intervals for real-robot object-goal navigation success rates in Table~\ref{tab:real_robot_results}. Values are percentages; each setting has $n=10$ trials.}
\label{tab:real-robot-ci}
\scriptsize
\setlength{\tabcolsep}{5pt}
\renewcommand{\arraystretch}{0.95}
\begin{tabular}{lcccc}
\toprule
Method & LC & OR & LC+OR & All settings \\
\midrule
M+S (ORB-SLAM3) &
$30.0\,[10.8,60.3]$ &
$30.0\,[10.8,60.3]$ &
$20.0\,[5.7,51.0]$ &
$26.7\,[14.2,44.4]$ \\
M+S (RTAB-Map) &
$40.0\,[16.8,68.7]$ &
$60.0\,[31.3,83.2]$ &
$30.0\,[10.8,60.3]$ &
$43.3\,[27.4,60.8]$ \\
\textbf{CROSS} &
$\mathbf{70.0\,[39.7,89.2]}$ &
$\mathbf{80.0\,[49.0,94.3]}$ &
$\mathbf{80.0\,[49.0,94.3]}$ &
$\mathbf{76.7\,[59.1,88.2]}$ \\
\bottomrule
\end{tabular}
\end{table}

The real-robot intervals are wide because each condition contains only ten trials. Consequently, we treat the real-robot experiment as evidence that the representation transfers to embodied object-goal navigation under realistic changes, while avoiding strong claims of fine-grained statistical separation between all methods and all conditions.

\subsection{Topo-Bench Evaluation}
\label{app:topo-bench}

We evaluate robustness to perceptual aliasing using \emph{Topo-Bench}~\cite{wang2025topo}, which categorizes perceptual aliasing scenarios into three regimes: \emph{Ambiguous + Positive} (A+P), \emph{Positive Only} (P.O.), and \emph{Ambiguous Only} (A.O.). A+P represents revisits to previously mapped regions where strong distractor matches closely resemble the true location; P.O. corresponds to revisits without highly similar distractors; and A.O. captures novel regions that spuriously resemble known locations. To quantify balanced performance across these regimes, Topo-Bench reports \emph{Balanced Localization Accuracy} (BLA), defined as the geometric mean of accuracies on A+P, P.O., and A.O., thereby penalizing methods with imbalanced behavior. Further details regarding these scenarios and metrics can be found in \cite{wang2025topo}.

Quantitative results are summarized in Table~\ref{tab:topo_bench}. All methods exhibit low accuracy in the challenging A+P regime, but our approach achieves the highest A+P performance, demonstrating the effectiveness of the proposed THF mechanism in resolving aliasing under appearance changes. At the same time, it maintains performance comparable to prior SOTA methods on P.O. and A.O., resulting in the best overall BLA. This indicates stronger and more balanced robustness across diverse aliasing scenarios.

\begin{figure}[t]
  \centering
  \setlength{\tabcolsep}{1pt} 
  \renewcommand{\arraystretch}{1.0}
  \begin{tabular}{@{}c c c c c c@{}}
    \includegraphics[width=0.162\textwidth]{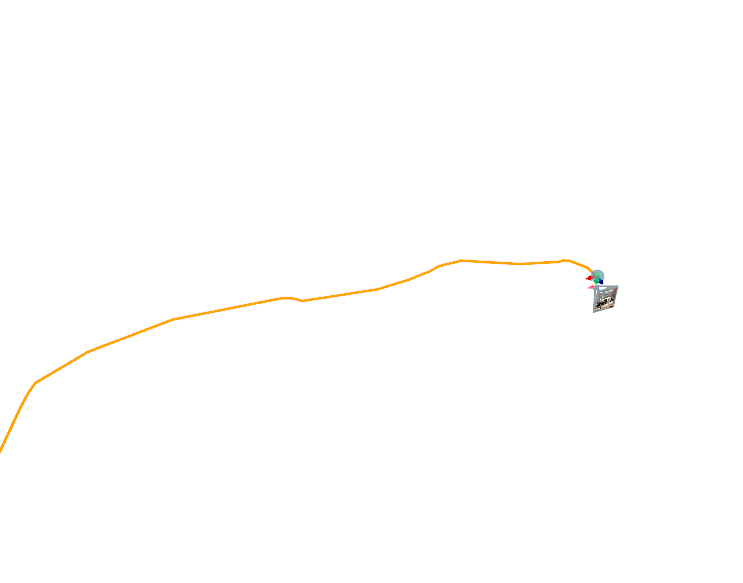} &
    \includegraphics[width=0.162\textwidth]{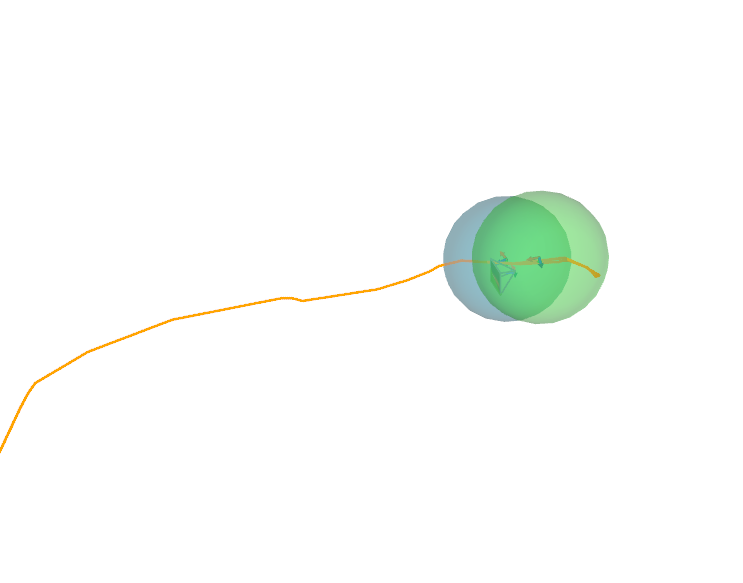} &
    \includegraphics[width=0.162\textwidth]{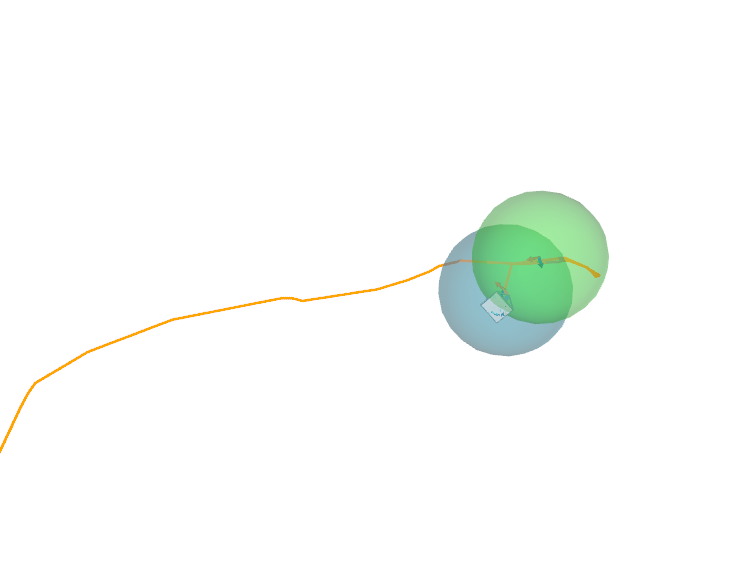} &
    \includegraphics[width=0.162\textwidth]{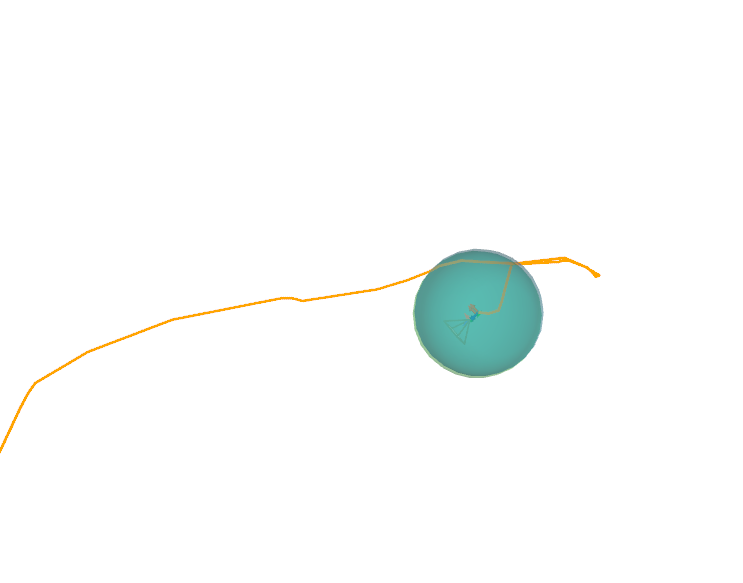} &
    \includegraphics[width=0.162\textwidth]{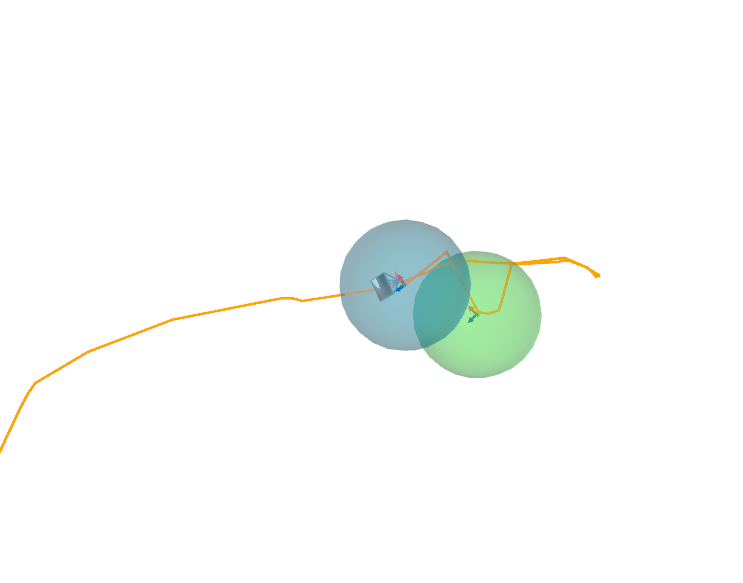} &
    \includegraphics[width=0.162\textwidth]{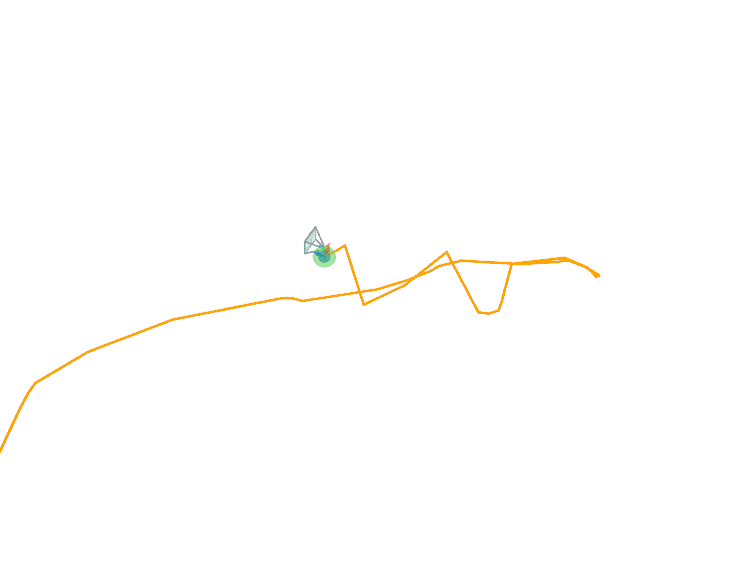} \\
    \includegraphics[width=0.162\textwidth]{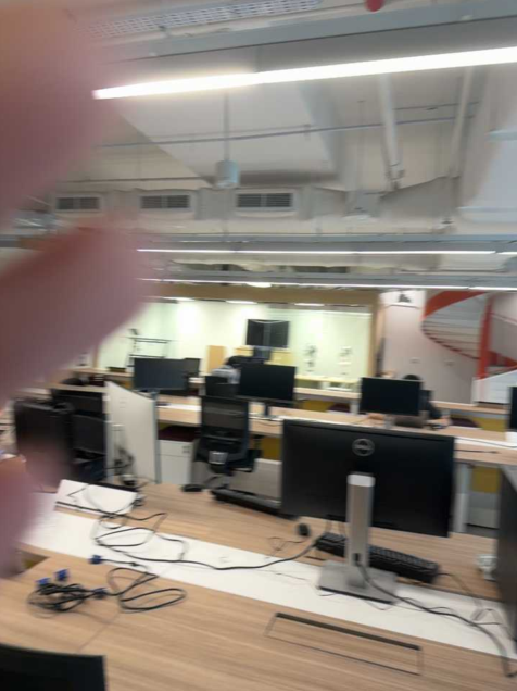} &
    \includegraphics[width=0.162\textwidth]{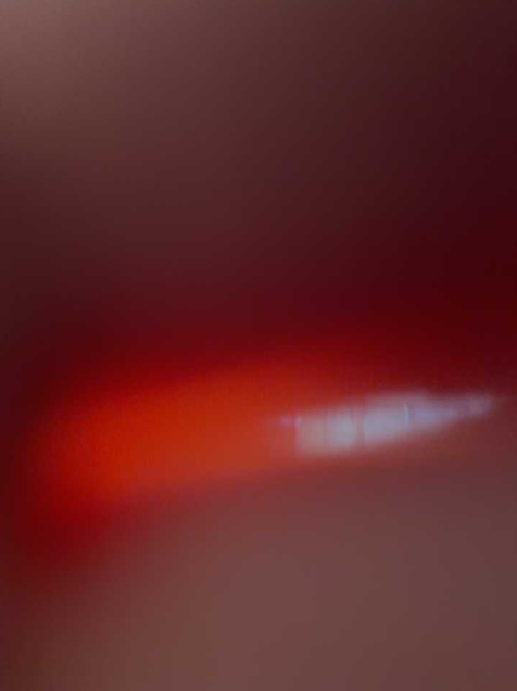} &
    \includegraphics[width=0.162\textwidth]{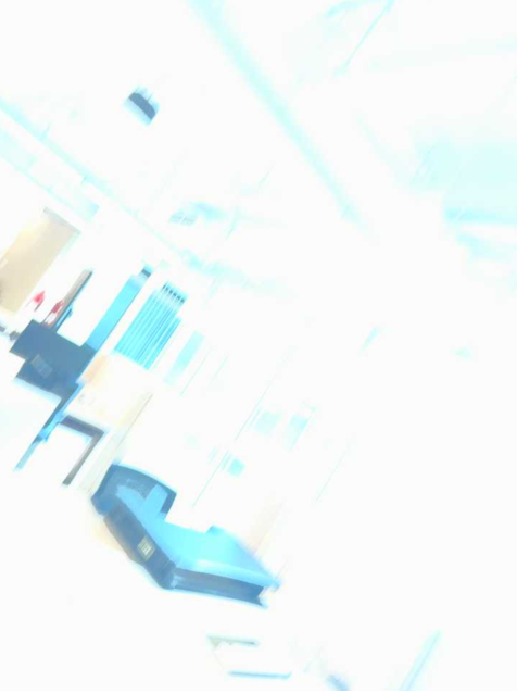} &
    \includegraphics[width=0.162\textwidth]{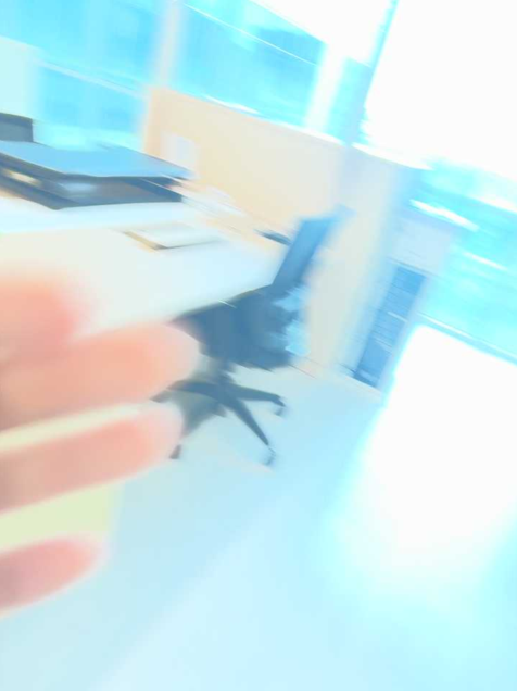} &
    \includegraphics[width=0.162\textwidth]{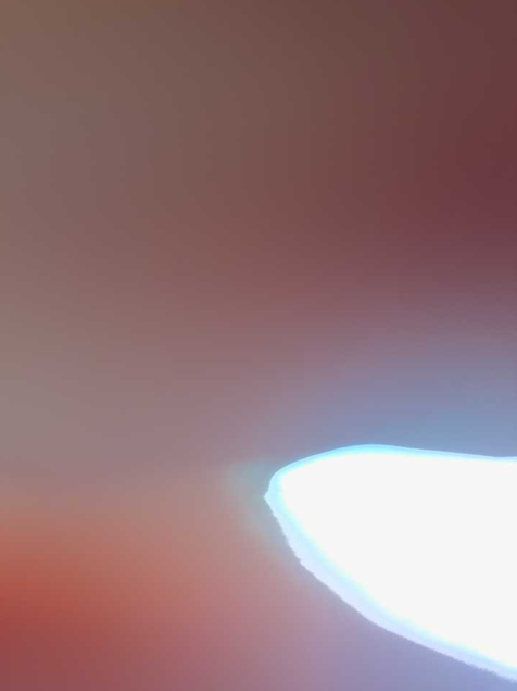} &
    \includegraphics[width=0.162\textwidth]{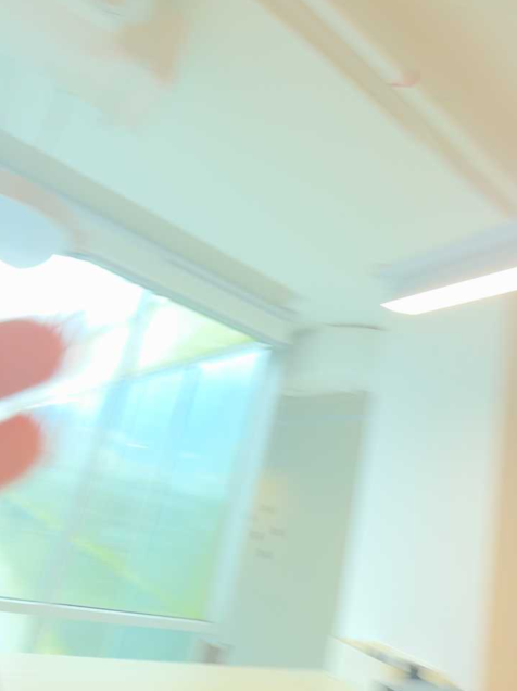} \\
    {\footnotesize(1)} & {\footnotesize(2)} & {\footnotesize(3)} &
    {\footnotesize(4)} & {\footnotesize(5)} & {\footnotesize(6)} \\
  \end{tabular}
    \caption{Occlusion sequence across six timestamps.
    Top row shows the online mapping and belief evolution of our system: the yellow line denotes the estimated trajectory; blue ellipses represent the propagated motion message, and green ellipses represent the measurement message. The lengths of the principal axes indicate the estimated uncertainty.
    Bottom row shows the corresponding RGB observations.
    At timestamp~(1), with no occlusion, uncertainty is low. At~(2) and~(3), occlusion prevents reliable visual observations, causing the motion message to continue propagating using local odometry while the measurement message remains static. At~(4), the two messages overlap, indicating successful retrieval and alignment with the stored map. At~(5), occlusion occurs again and uncertainty increases as the motion message propagates. Finally, at~(6), once the occlusion disappears, uncertainty rapidly decreases and the system localizes with high confidence.}

  \label{fig:occlusion_sequence}
\end{figure}

\subsection{Robustness to Odometry Noise, Occlusion, and Fast Motion}
\label{app:robustness-others}
We found our system is robust to fast motion and camera occlusion: the states are propagated when image matching is not possible due to the camera blur or occlusion, but the global measurement message using VPR model can quickly retrieve the correct frame once it is available. 

\begin{figure}[t]
  \centering
  \setlength{\tabcolsep}{1pt} 
  \renewcommand{\arraystretch}{1.0}
  \begin{tabular}{@{}c c c c c c c@{}}
    \includegraphics[width=0.14\textwidth]{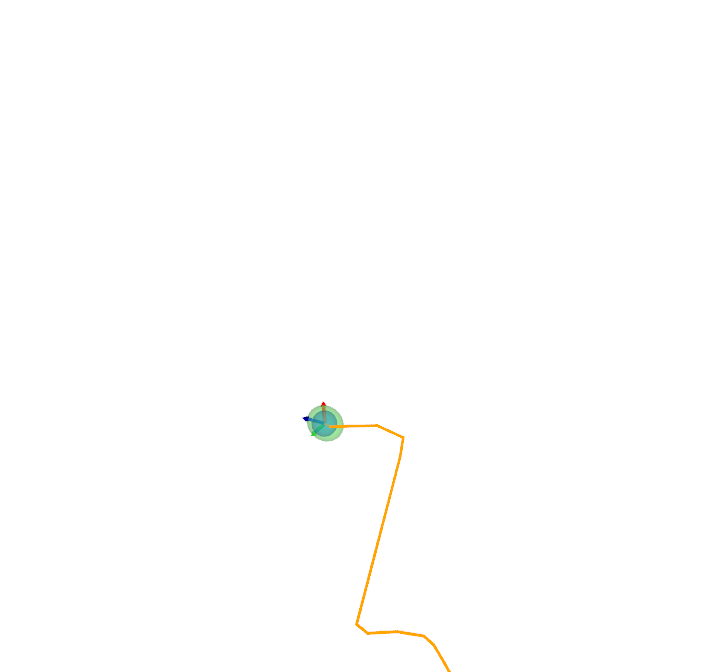} &
    \includegraphics[width=0.14\textwidth]{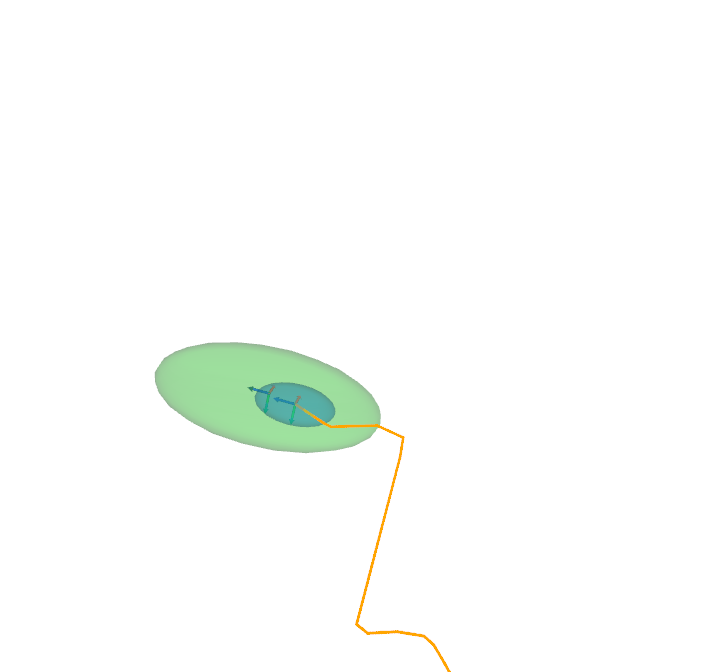} &
    \includegraphics[width=0.14\textwidth]{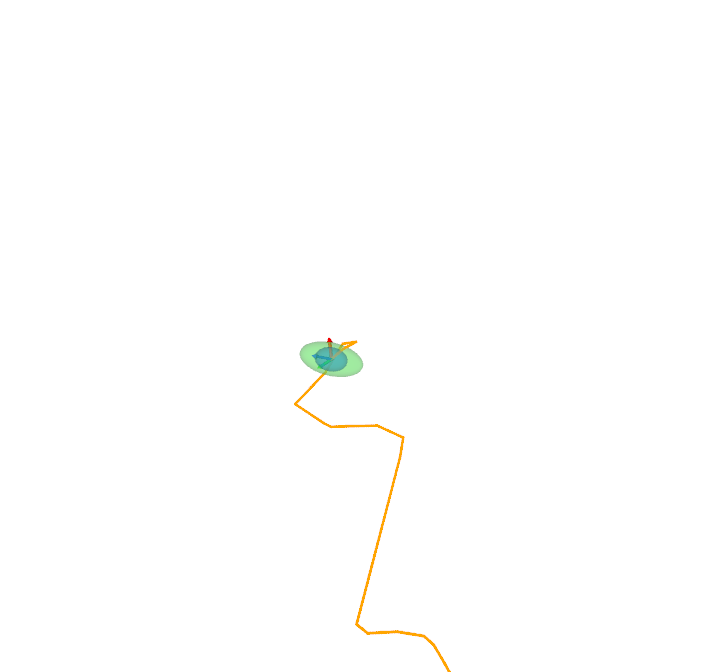} &
    \includegraphics[width=0.14\textwidth]{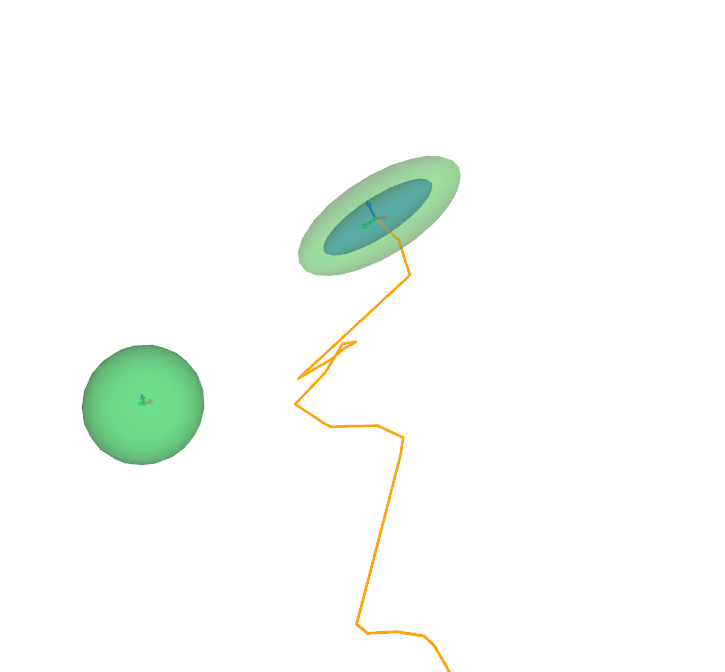} &
    \includegraphics[width=0.14\textwidth]{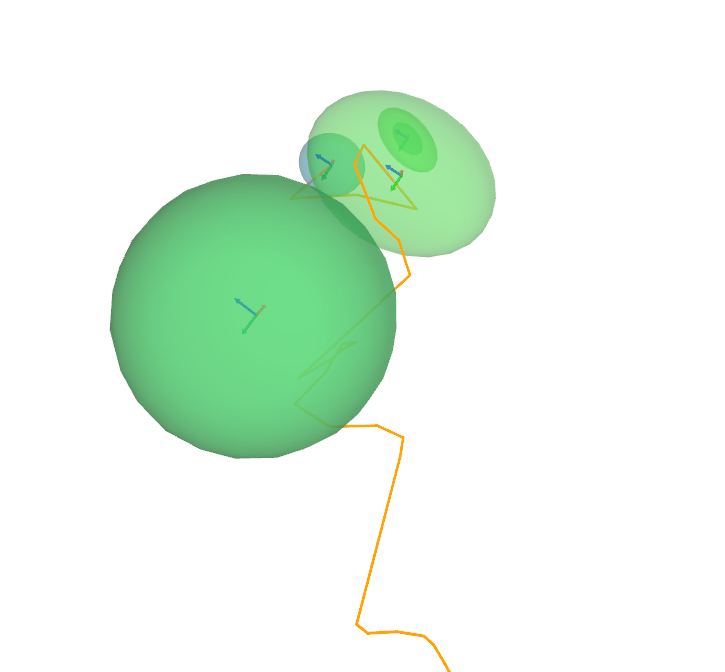} &
    \includegraphics[width=0.14\textwidth]{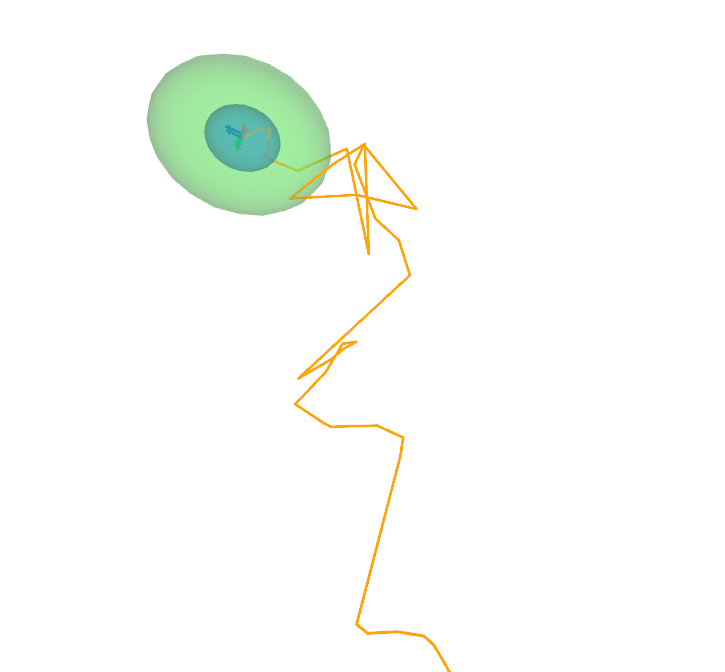} &
    \includegraphics[width=0.14\textwidth]{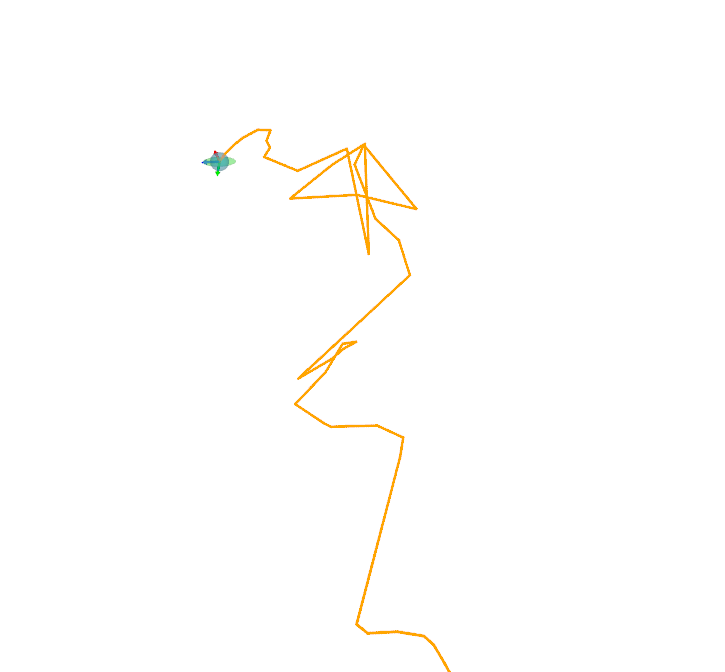} \\
    \includegraphics[width=0.14\textwidth]{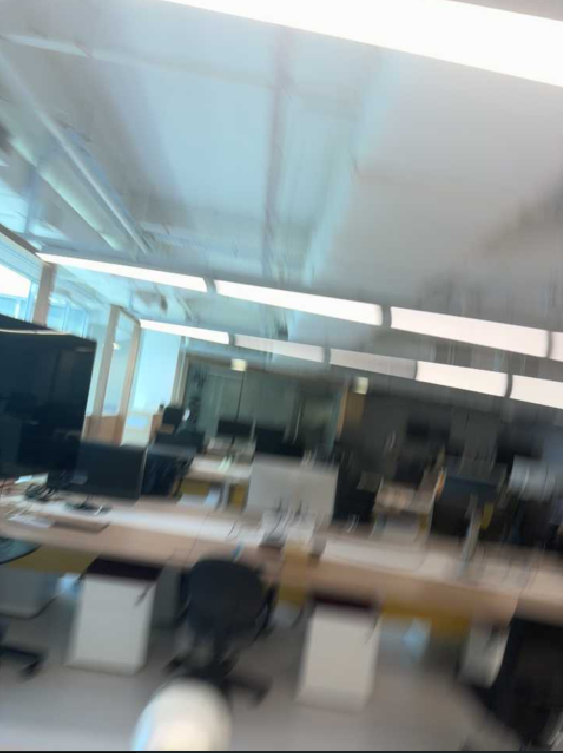} &
    \includegraphics[width=0.14\textwidth]{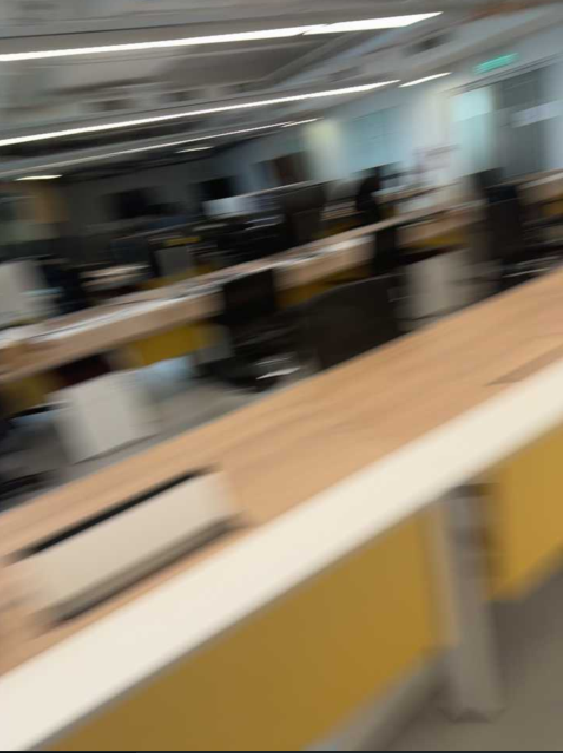} &
    \includegraphics[width=0.14\textwidth]{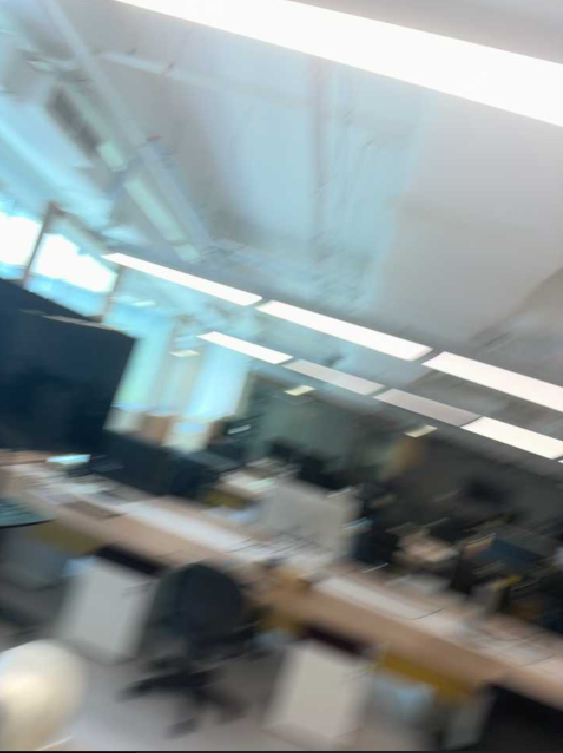} &
    \includegraphics[width=0.14\textwidth]{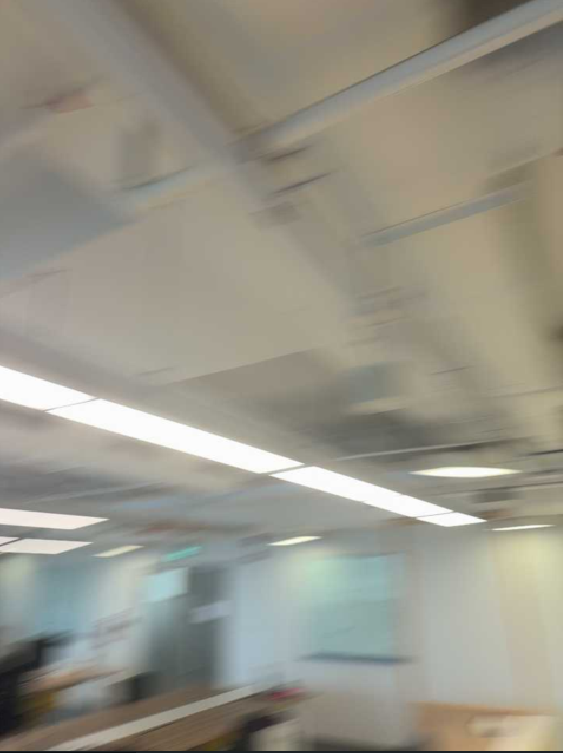} &
    \includegraphics[width=0.14\textwidth]{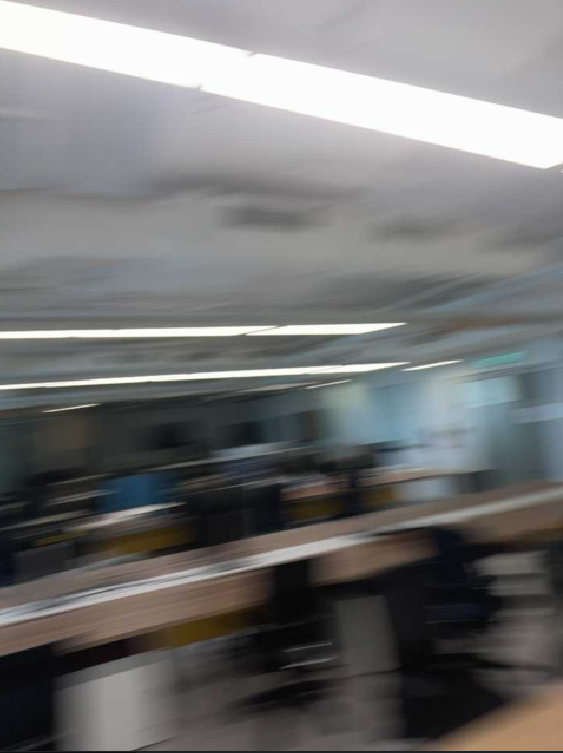} &
    \includegraphics[width=0.14\textwidth]{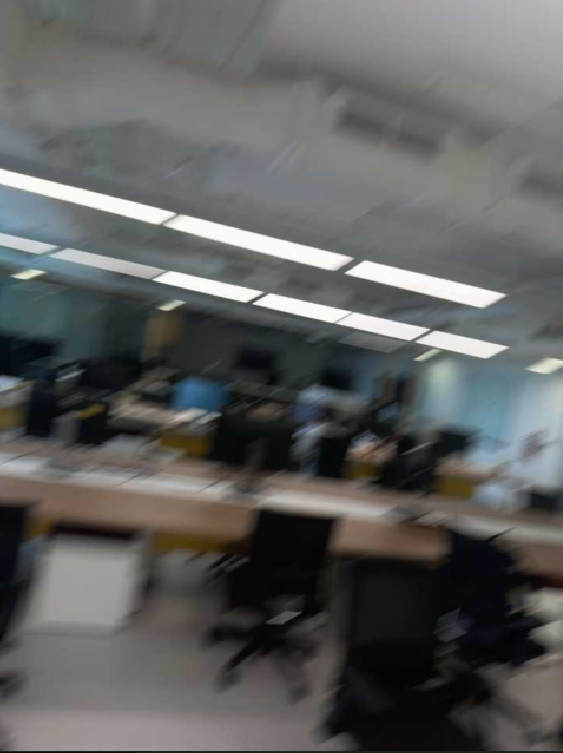} &
    \includegraphics[width=0.14\textwidth]{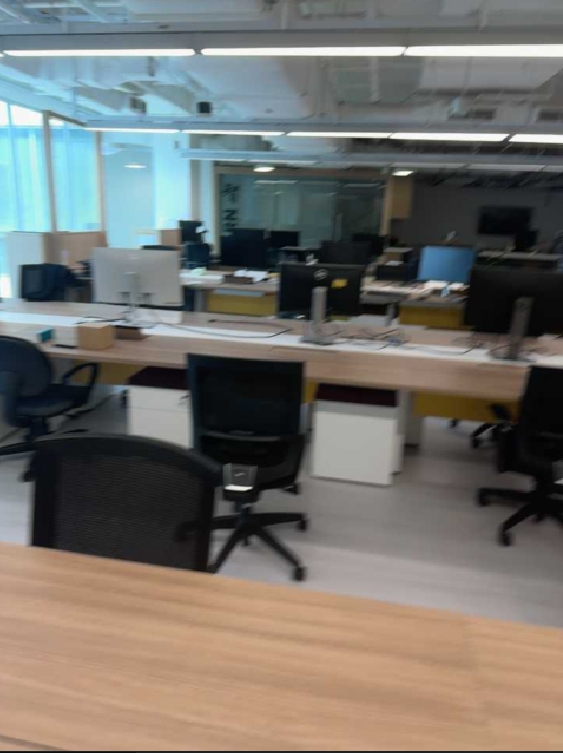} \\
    {\footnotesize(1)} & {\footnotesize(2)} & {\footnotesize(3)} &
    {\footnotesize(4)} & {\footnotesize(5)} & {\footnotesize(6)} & {\footnotesize(7)} \\
  \end{tabular}
    \caption{Fast-motion sequence across seven timestamps. 
    Top row shows the online mapping trajectory of our system: yellow lines indicate the estimated trajectory; blue ellipses denote the Gaussian mixture of the measurement message, while cyan ellipses denote the Gaussian mixture of the propagated motion message. The principal axis lengths of each ellipse correspond to the estimated uncertainty (variance) along that direction. 
    Bottom row shows the corresponding RGB observations. 
    At timestamp~(1), uncertainty is low. At~(2), observation uncertainty increases significantly due to severe motion blur. At~(3), the system maintains tracking with high uncertainty despite residual blur. At~(4) and~(5), observation uncertainty becomes very large again due to heavy blur, introducing spurious hypotheses. At~(6) and~(7), the system rapidly eliminates these spurious hypotheses via temporal hypothesis filtering once visual observations return to normal.}
  \label{fig:fast_motion}
\end{figure}

Figure~\ref{fig:fast_motion} and Figure~\ref{fig:occlusion_sequence} provide qualitative demonstrations of how the proposed system performs when fast motion and occlusion occur.

\begin{figure}[t]
  \centering
  \setlength{\tabcolsep}{2pt}
  \renewcommand{\arraystretch}{1.0}
  \begin{tabular}{c c c}

    \includegraphics[width=0.329\textwidth]{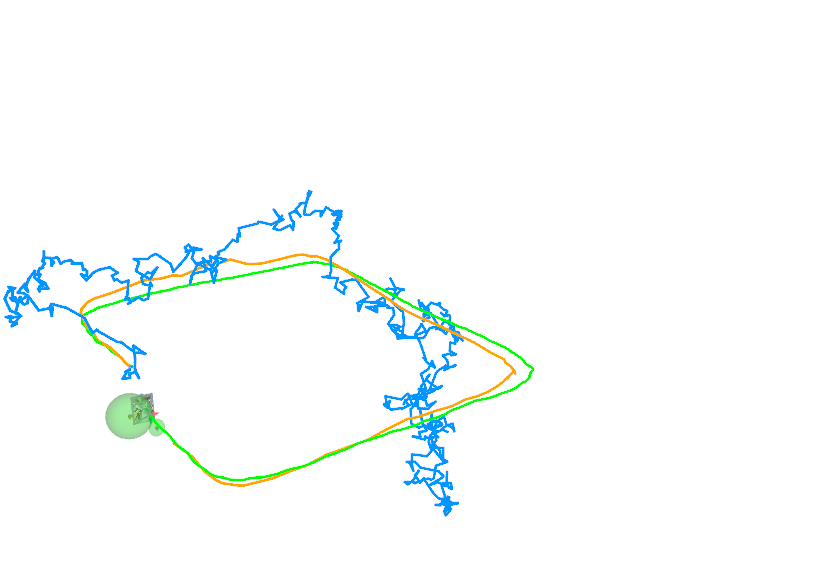} &
    \includegraphics[width=0.329\textwidth]{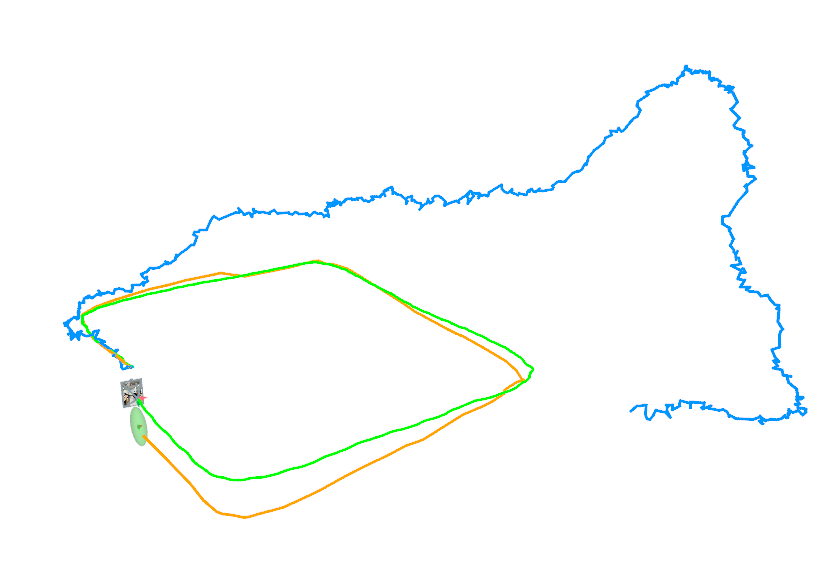} &
        \includegraphics[width=0.329\textwidth]{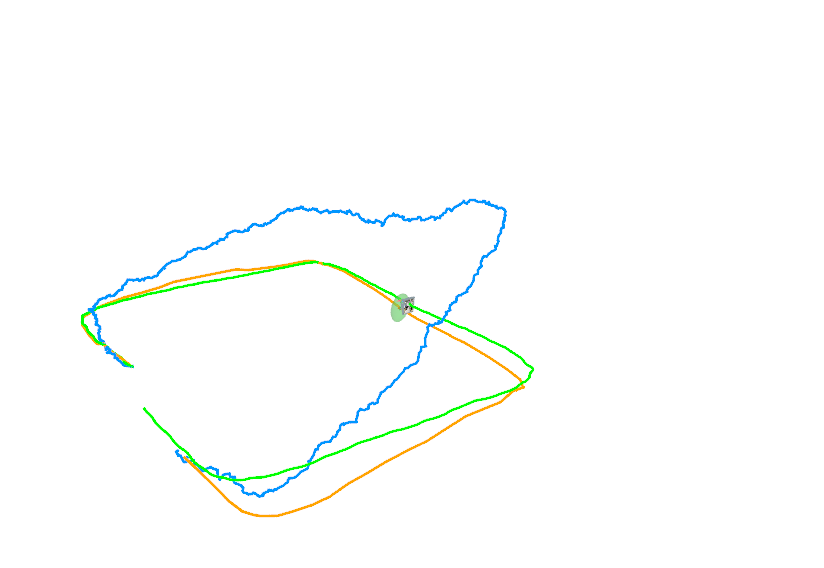}\\
    {\footnotesize SNR=0.2} &
    {\footnotesize SNR=0.5} &
    {\footnotesize SNR=1}
  \end{tabular}
    \caption{Noisy-odometry experiment under different signal-to-noise ratio (SNR) settings.
    Each column shows the full trajectory for the same sequence with increasing levels of odometry noise injected via right-multiplicative perturbations in $\mathrm{SE}(3)$.
    The yellow line denotes the trajectory estimated by our system, the green line shows the reference trajectory provided by the SLAM system, and the blue line corresponds to the corrupted odometry used for motion propagation.
    As the SNR decreases, the odometry becomes increasingly distorted, as evidenced by the highly corrupted blue trajectories.
    Notably, even at SNR$=0.2$, which corresponds to very large noise relative to the true odometry measurements, our estimated trajectory remains close to the reference, demonstrating robustness to severe odometric corruption.}
    \label{fig:noisy_odometry}

\end{figure}

Our system replies on the odometry information for high frequency updates. To test how our system performs when the odometry is noisy, we record trajectories with RGBD and pose information using the ARKit API on iPhone. We then add noise to the computed delta pose between two consecutive frames as the noisy odometry input as follows:

Given an odometry pose $T\in\mathrm{SE}(3)$, we apply a right-multiplicative perturbation
\[
\tilde{T} = T\,\exp(\xi), \qquad 
\xi=\begin{bmatrix}\delta\rho\\ \delta\phi\end{bmatrix}\in\mathbb{R}^6,
\]
where $\exp(\cdot)$ is the Lie exponential. The translational and rotational noises are sampled as
\[
\delta\rho \sim \mathcal{N}(0,\sigma_t^2 I_3), \qquad
\delta\phi \sim \mathcal{N}(0,\sigma_r^2 I_3),
\]
with units of meters and radians, respectively.

To obtain a scale-aware and interpretable noise level, we set
\[
\sigma_t = \frac{\|t\|}{\mathrm{snr}\sqrt{3}}, \qquad
\sigma_r = \frac{\|\log(R)\|}{\mathrm{snr}\sqrt{3}},
\]
where $T=\begin{bmatrix}R&t\\0&1\end{bmatrix}$ and $\log(R)\in\mathfrak{so}(3)$. 
Since $\mathbb{E}\|\delta\rho\|^2=3\sigma_t^2$ and $\mathbb{E}\|\delta\phi\|^2=3\sigma_r^2$, the expected noise magnitudes scale as $\|t\|/\mathrm{snr}$ and $\|\log(R)\|/\mathrm{snr}$, respectively, yielding an SNR-like control over pose perturbations.

Figure~\ref{fig:noisy_odometry} provides a qualitative evaluation of our system under different odometry noise levels. 
As the signal-to-noise ratio (SNR) decreases, the injected perturbations induce progressively larger drift in the motion propagation. 
Despite increasingly noisy odometry, the system remains stable and is able to recover and maintain a coherent trajectory by leveraging visual observations and temporal hypothesis filtering, demonstrating robustness to substantial odometric corruption.

\subsection{Ablation Study}
\label{subsec:ablation}

Our ablations are designed to diagnose the two components most directly tied to the central claim: 
(i) representing competing explanations as continuous $\mathrm{SE}(3)$ pose modes rather than discrete place IDs, and 
(ii) constructing global measurement modes using geometrically verified relative pose estimates.

\mypara{Continuous versus discrete hypothesis testing.}
PBU is the closest controlled baseline for testing the state representation. 
It uses the same topological graph, the same BoQ retrieval scores, the same query protocol, and the same high-precision calibration procedure, but its belief is defined over discrete node IDs rather than continuous poses. 
Thus, the PBU--CROSS comparison isolates the additional value of retaining continuous pose evidence during temporal hypothesis filtering. 
As shown in Table~\ref{tab:appearance-change}, CROSS improves aggregate cross-sequence relocalization success from $0.068$ to $0.384$ over PBU, with the largest gain on Rover ($0.063$ to $0.397$). 
This indicates that many visually plausible matches cannot be rejected by discrete temporal consistency alone; maintaining hypotheses in $\mathrm{SE}(3)$ allows odometry to test whether a retrieved place is physically consistent before promoting it.

\mypara{Beyond VPR-only temporal aggregation.}
GM, SM, and PBU share the same VPR backend where applicable and differ primarily in how temporal evidence is accumulated: single-frame retrieval, sequence-level aggregation, and discrete Bayesian filtering. 
ABM provides an additional appearance-based mapping baseline. 
The lower cross-sequence success rates of these methods indicate that CROSS's improvement is not due solely to the BoQ descriptor or to temporal smoothing of retrieval scores. 
Instead, the gain comes from coupling VPR proposals with relative-pose verification and delayed branch validation in continuous pose space.

\mypara{Relative-pose estimator.}
Another key design choice is the use of a classical PnP-based pipeline for relative pose estimation between frames. 
As an alternative, we evaluated a learned variant by replacing the XFeat/LightGlue + PnP module with VGGT~\cite{wang2025vggt}. 
In our relocalization setting, this variant achieved zero success rate. 
We observed two practical failure modes: VGGT predicts poses with arbitrary scale across independent inferences, which prevents reliable global pose composition, and its average runtime is $490$\,ms per forward pass on an RTX4000 GPU, compared with $16$\,ms for the PnP-based relative-pose module. 
This ablation does not imply that learned relative-pose estimation is unsuitable in principle; rather, it shows that the current system benefits from a metric RGB-D PnP estimate whose scale is fixed and whose inlier support provides an explicit quality signal for measurement weighting.

\begin{figure}[t]
    \centering
    \includegraphics[width=0.7\linewidth]{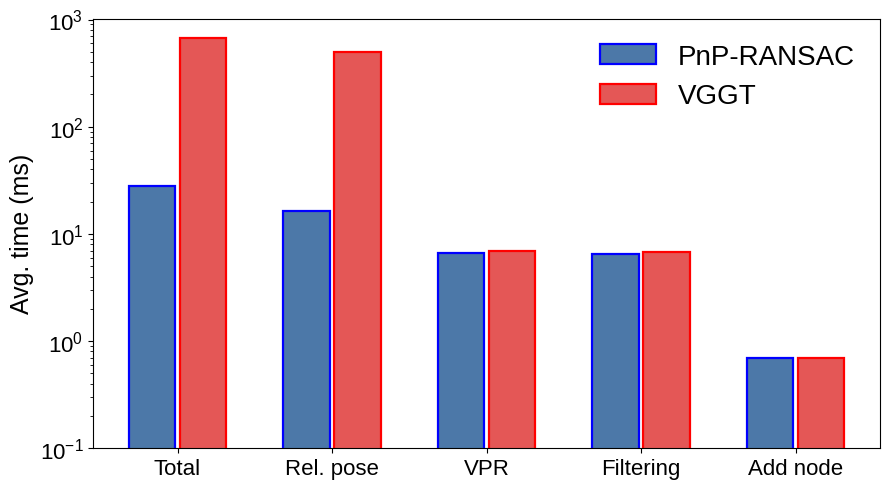}
    \caption{Runtime breakdown of the mapping pipeline per step, comparing relative pose estimation via \textbf{PnP-RANSAC} versus \textbf{VGGT}~\cite{wang2025vggt}. Bars report the \emph{total} average step time and its main components: relative pose estimation (Rel.\ pose), visual place recognition retrieval (VPR), and temporal hypothesis filtering. The log-scale y-axis highlights the large disparity in pose-estimation cost.}
    \label{fig:runtime}
\end{figure}

\subsection{Runtime Analysis}
\label{subsec:runtime}

Figure~\ref{fig:runtime} presents a runtime breakdown of our system measured on an RTX~4000 GPU workstation, detailing the computation time of each major module. The complete pipeline requires approximately 28\,ms per step, enabling real-time operation at over 30\,Hz.

In practical deployments with more limited computational resources, the full pipeline need not be executed at such a high frequency. Local odometry can propagate the robot state at high rate between global updates, while the observation module can construct observations at a lower frequency without degrading overall performance.

\subsection{Compute Resources}
\label{app:compute-resources}

Offline experiments were run on two local workstations; no cloud TPU/large-scale
cluster resources were used. The main GPU workstation used an AMD Ryzen 9 9900X
(12 physical cores, 24 threads), 59.4~GB system memory, an NVIDIA RTX 4000 Ada
Generation GPU with 19.6~GB VRAM, a 937~GB NVMe system SSD, and a 4~TB NVMe data
SSD. RTAB-Map was run CPU-only on a separate AMD Ryzen Threadripper 9960X
workstation with 48 logical threads and 128~GB system memory. The software
environment used Linux 6.11.0, CUDA 12.6, Python 3.11.13, PyTorch
2.7.1+cu126, torchvision 0.22.1+cu126, OpenCV 4.11.0, NumPy 1.26.4, and
pandas 2.3.1.

For CROSS, the online mapping/localization pipeline runs in approximately
28~ms per RGB-D frame on an RTX 4000 GPU workstation, including visual place
recognition, relative pose estimation, and temporal hypothesis filtering
(Fig.~\ref{fig:runtime}). The PnP-based relative pose module takes
approximately 16~ms per frame; the VGGT ablation takes approximately
490~ms per forward pass on the same GPU, and was therefore not used in the
main real-time system.

The real robot object navigation experiment is conducted on a Unitree Go2 EDU quadruped robot platform, equipped with an Intel Realsense D435i camera. The robot's onboard computer is connected to an off-board HP Omen Transcend 14 Laptop (Intel Core Ultra 7 155H processor, 16 GB RAM, NVIDIA GeForce RTX 4050 GPU) over WiFi. The camera data is streamed to the off-board laptop for updating the CROSS map and pose estimate in real time. This off-board setup is used to enable faster compute cycles as well as a longer runtime for the robot; as against an on-board processing unit which would either draw power from the robot's battery or an external mounted battery, which would in turn add payload weight decreasing the robot's runtime.

Table~\ref{tab:compute-resources} summarizes the compute needed to reproduce
each reported experiment. Wall-clock times include mapping, query-time
localization, baseline evaluation, metric computation, and logging, but exclude
dataset download time. Storage includes the raw datasets, extracted features,
intermediate logs, and output result files. The table separates the
OpenLORIS/Rover CROSS benchmark runs, the corresponding SLAM/topological
baseline runs, the Topo-Bench evaluation, and the robot/ablation experiments.
The full research project did
not require substantially more compute than the experiments
reported here; exploratory development and failed runs required approximately
100 additional hours.

\begin{table}[t]
\centering
\caption{Compute resources required to reproduce the experiments. 
\textbf{Offline experiments} (OpenLORIS, Rover, Topo-Bench) were conducted on an RTX 4000 workstation (Ryzen 9 9900X, 60GB RAM). 
\textbf{Real-robot experiments} utilized a Unitree Go2 EDU robot equipped with a Realsense D435i camera, with the computations on an offboard HP Omen Transcend 14 laptop (Intel Core 7 155H, RTX 4050 6GB VRAM). Real-robot compute numbers also include the compute required to run the object search algorithm to find the target frame for object-goal navigation, which contributes to VRAM usage for the other baseline methods.}
\label{tab:compute-resources}
\scriptsize
\setlength{\tabcolsep}{4pt}
\renewcommand{\arraystretch}{1.2}
\begin{tabular}{p{0.22\linewidth} p{0.20\linewidth} p{0.22\linewidth} p{0.18\linewidth}}
\toprule
Method & Time / step & Total wall-clock time & Peak RAM/VRAM \\
\midrule
\multicolumn{4}{@{}l}{\textit{OpenLORIS Corridor + Rover Campus (Total: 735 + 1232 query trials)}} \\[0.25em]
CROSS & 28~ms/frame & $\approx$139~hours & $\approx$35/2~GB \\
RTAB-Map & 36~ms/frame & $\approx$170~hours & $\approx$110/0~GB \\
ORB-SLAM3 & 14~ms/frame & $\approx$10~hours & $\approx$6/0~GB \\
MASt3R-SLAM & 74~ms/frame & $\approx$52~hours & $\approx$32/18~GB \\
ABM & 17~ms/frame & $\approx$85~hours & $\approx$28/0~GB \\
GM / SM / PBU & 19~ms/frame & $\approx$90~hours & $\approx$5/1~GB \\

\midrule
\multicolumn{4}{@{}l}{\textit{Topo-Bench perceptual-aliasing (Total: 513 query trials)}} \\[0.25em]
CROSS & 28~ms/frame & $\approx$20~hours & $\approx$36/2~GB \\
RTAB-Map & 36~ms/frame & $\approx$26~hours & $\approx$110/0~GB \\
ORB-SLAM3 & 14~ms/frame & $\approx$2~hours & $\approx$6/0~GB \\
ABM & 17~ms/frame & $\approx$12~hours & $\approx$28/0~GB \\
OpenFabmap & 130~ms/frame & $\approx$90~hours & $\approx$38/0~GB \\
Rat-Slam & 85~ms/frame & $\approx$61~hours & $\approx$29/0~GB \\
GM / SM / PBU & 19~ms/frame & $\approx$14~hours & $\approx$5/1~GB \\

\midrule
\multicolumn{4}{@{}l}{\textit{Real-robot object navigation (3 settings $\times$ 10 trials)}} \\[0.25em]
CROSS & 120~ms/frame & $\approx$1~hour & $\approx$3.5/4.5~GB \\
Metric-Semantic (RTAB-MAP) & 90~ms/frame & $\approx$1~hour & $\approx$1.3/4.5~GB \\
Metric-Semantic (ORBSLAM) & 34~ms/frame & $\approx$1~hour & $\approx$2.1/4.5~GB \\
\bottomrule
\end{tabular}
\end{table}

\section{Broader Impacts}
\label{app:broader-impacts}

CROSS is intended to improve long-term robot navigation by enabling robots to reuse spatial memories under illumination, seasonal, object-level, and dynamic-scene changes. Potential positive impacts include more reliable service, inspection, delivery, and assistive robots, especially in environments where frequent remapping is costly or impractical. The sparse topological representation may also reduce storage and computation relative to dense metric-semantic maps, which can make long-term semantic navigation more accessible on resource-constrained robotic platforms.

The same capabilities also introduce potential risks. A system that stores RGB-D observations for later semantic querying may capture people, private spaces, or sensitive objects, raising privacy concerns if deployed without consent, retention limits, or access control. More robust relocalization and navigation could also be misused for unauthorized surveillance, tracking, or physical intrusion. Finally, navigation failures caused by incorrect relocalization, stale observations, biased semantic detections, or outdated object memories could cause a robot to navigate to an unintended location or act on scene information that no longer reflects the current environment.

Several mitigation measures are appropriate for deployment. Stored visual memories should be minimized to task-relevant environments, protected with access controls, and deleted or anonymized when no longer needed. Deployments should obtain appropriate consent in private or semi-private spaces and should avoid using the system for identifying, tracking, or inferring sensitive attributes of people. Safety-critical applications should include conservative navigation policies, obstacle avoidance independent of the global memory, confidence thresholds for relocalization, and human override or monitoring when operating near people. The present work is an enabling navigation method and does not release a high-risk generative model or scraped dataset, but practical use should still follow local privacy, robotics safety, and data-governance requirements.

\section{Additional Method Details}
\label{app:method-details}
\subsection{Classical Gaussian--sum Filtering (GSF)}
\label{app:gsf}

This appendix briefly summarizes the classical Gaussian--sum filter (GSF).
We first present the exact linear--Gaussian case, then the standard nonlinear
extension via per-component local Gaussianization (EKF/UKF/CKF). All derivations
are written in a local Euclidean chart; on Lie groups (e.g., $\SE(3)$) the same
updates are applied in a consistent tangent chart.

\subsubsection{Preliminaries and Notation}
\label{app:GSF:prelim}

Let $x_t\in\R^n$ be the (locally Euclidean) state with motion and measurement
models
\begin{equation}
\begin{aligned}
x_t &= f_t(x_{t-1},u_{t-1}) + q_t, \qquad q_t \sim \Normal(0,Q_t),\\
z_t &= h_t(x_t) + r_t, \qquad\quad\;\;\; r_t \sim \Normal(0,R_t),
\end{aligned}
\label{eq:app:models}
\end{equation}
where $Q_t,R_t \succ 0$. The GSF represents the filtering density as a finite
mixture
\begin{equation}
p(x_{t-1}\!\mid z_{1:t-1},u_{1:t-2})
= \sum_{k=1}^{K_{t-1}} w^{(k)}_{t-1}\,
\Normal\!\big(x_{t-1};\mu^{(k)}_{t-1},\Sigma^{(k)}_{t-1}\big),
\nonumber
\end{equation}
with $w^{(k)}_{t-1}\!\ge\!0$ and $\sum_k w^{(k)}_{t-1}\!=\!1$.
On a Lie group $\mathcal{X}$, one works in a fixed local chart
$\phi:\mathcal{X}\!\to\!\R^n$ (e.g., left-/right-invariant log), performs the
Euclidean update on $\xi_t=\phi(x_t)$, and maps back via the exponential.

\subsubsection{Exact Linear--Gaussian GSF}

Assume linear--Gaussian models:
\begin{equation}
x_t = F_t x_{t-1} + B_t u_{t-1} + q_t,
\qquad
z_t = H_t x_t + r_t .
\nonumber
\end{equation}

\paragraph{Prediction (per component).}
For each $k=1,\dots,K_{t-1}$,
\begin{equation}
\begin{aligned}
\mu^{(k)}_{t\mid t-1} &= F_t \mu^{(k)}_{t-1} + B_t u_{t-1},\\
\Sigma^{(k)}_{t\mid t-1} &= F_t \Sigma^{(k)}_{t-1} F_t^\top + Q_t,\\
w^{(k)}_{t\mid t-1} &= w^{(k)}_{t-1}.
\end{aligned}
\label{eq:app:pred}
\end{equation}
Thus
$p(x_t\!\mid z_{1:t-1},u_{1:t-1})
=\sum_k w^{(k)}_{t\mid t-1}\Normal(x_t;\mu^{(k)}_{t\mid t-1},\Sigma^{(k)}_{t\mid t-1})$.

\paragraph{Update (per component Kalman update).}
Define, for each component,
\begin{equation}
\begin{aligned}
y^{(k)}_t &= z_t - H_t \mu^{(k)}_{t\mid t-1},\\
S^{(k)}_t &= H_t \Sigma^{(k)}_{t\mid t-1} H_t^\top + R_t,\\
K^{(k)}_t &= \Sigma^{(k)}_{t\mid t-1} H_t^\top \big(S^{(k)}_t\big)^{-1}.
\end{aligned}
\label{eq:app:innov}
\end{equation}
Then
\begin{equation}
\begin{aligned}
\mu^{(k)}_{t} &= \mu^{(k)}_{t\mid t-1} + K^{(k)}_t\, y^{(k)}_t,\\
\Sigma^{(k)}_{t} &= \big(I - K^{(k)}_t H_t\big)\Sigma^{(k)}_{t\mid t-1}.
\end{aligned}
\label{eq:app:post}
\end{equation}
Weights update by the marginal likelihood (Kalman evidence):
\begin{equation}
\tilde w^{(k)}_{t}
= w^{(k)}_{t\mid t-1}\,\Normal\!\big(y^{(k)}_t;0,S^{(k)}_t\big),
\qquad
w^{(k)}_t=\frac{\tilde w^{(k)}_t}{\sum_j \tilde w^{(j)}_t}.
\label{eq:app:weight}
\end{equation}
The posterior remains a mixture
$p(x_t\!\mid z_{1:t})=\sum_k w^{(k)}_t \Normal(x_t;\mu^{(k)}_t,\Sigma^{(k)}_t)$.

\subsubsection{Nonlinear GSF via Local Gaussianization}
\label{app:GSF:nonlin}

For nonlinear \eqref{eq:app:models}, GSF applies a local Gaussian filter to
each component.

\paragraph{EKF-style (per component).}
Linearize around the current component mean:
\begin{equation}
\begin{aligned}
f_t(x,u) &\approx f_t(\mu^{(k)}_{t-1},u) + F^{(k)}_t (x-\mu^{(k)}_{t-1}),\\
h_t(x) &\approx h_t(\mu^{(k)}_{t\mid t-1}) + H^{(k)}_t (x-\mu^{(k)}_{t\mid t-1}),
\end{aligned}
\nonumber
\end{equation}
where $F^{(k)}_t,H^{(k)}_t$ are Jacobians. Then apply
\eqref{eq:app:pred}--\eqref{eq:app:weight} with $(F_t,H_t)$ replaced by
$(F^{(k)}_t,H^{(k)}_t)$ and with the corresponding affine terms.

\paragraph{UKF/CKF (per component).}
Alternatively, propagate sigma points/quadrature points per component to obtain
$(\mu^{(k)}_{t\mid t-1},\Sigma^{(k)}_{t\mid t-1})$ and predicted measurement
moments, then use \eqref{eq:app:innov}--\eqref{eq:app:weight}.

\subsubsection{Mixture Identities}
\label{app:GSF:idents}

Let $\Normal_i(x)=\Normal(x;m_i,S_i)$ for $i\in\{1,2\}$.

\paragraph{Product of Gaussians.}
\begin{equation}
\Normal_1(x)\Normal_2(x)
=
\Normal(m_1;m_2,S_1{+}S_2)\;
\Normal(x;m,S),
\label{eq:app:prod}
\end{equation}
where $S=(S_1^{-1}+S_2^{-1})^{-1}$ and $m=S(S_1^{-1}m_1+S_2^{-1}m_2)$.

\paragraph{Innovation evidence.}
For predicted $(\mu^-,\Sigma^-)$ and measurement $z=Hx+r$, $r\!\sim\!\Normal(0,R)$,
the innovation $y=z-H\mu^-$ satisfies
$y\sim \Normal(0,S)$ with $S=H\Sigma^-H^\top+R$, yielding the evidence term in
\eqref{eq:app:weight}.

\subsubsection{Mixture Growth Control}

To prevent unbounded mixture growth, GSF typically uses:

\paragraph{Pruning.}
Remove components with $w^{(k)}_t < \varepsilon$.

\paragraph{Reduction / merging.}
Iteratively merge nearby components (e.g., using a KL-based criterion) until
$K_t\le K_{\max}$. Merging two components with weights $a,b$ by moment matching
gives
\begin{equation}
\mu = \frac{a\mu_a+b\mu_b}{a+b},
\label{eq:app:mm-mean}
\end{equation}

\begin{equation}
\small
\Sigma =
\frac{a\!\left(\Sigma_a + (\mu_a-\mu)(\mu_a-\mu)^\top\right)
      +b\!\left(\Sigma_b + (\mu_b-\mu)(\mu_b-\mu)^\top\right)}{a+b}.
\nonumber
\end{equation}

\paragraph{Splitting (optional).}
If a component violates a consistency/nonlinearity test, split it into a small
set whose moments match the parent, and distribute the parent weight across the
children.

\subsubsection{Mixture--Mixture Update (Optional)}

If the measurement factor is approximated by a mixture
$Q_t(x)=\sum_{c=1}^{C_t}\pi^{(c)}_t \Normal(x;\nu^{(c)}_t,\Lambda^{(c)}_t)$,
then the update is a mixture--mixture product:
\begin{equation}
p(x_t\!\mid z_{1:t}) \propto p^-(x_t)\,Q_t(x_t),
\nonumber
\end{equation}
with
\begin{equation}
p^-(x_t)=\sum_{k} w^{(k)}_{t\mid t-1}
\Normal\!\big(x_t;\mu^{(k)}_{t\mid t-1},\Sigma^{(k)}_{t\mid t-1}\big),
\nonumber
\end{equation}
and
\begin{equation}
p(x_t\!\mid z_{1:t})
=
\sum_{k=1}^{K_{t-1}}\sum_{c=1}^{C_t}
\tilde w_{k,c}\;\Normal(x_t; m_{k,c}, S_{k,c}).
\label{eq:app:mm-update}
\end{equation}
Here $(m_{k,c},S_{k,c})$ follow \eqref{eq:app:prod} and
\begin{equation}
\tilde w_{k,c}
\propto
w^{(k)}_{t\mid t-1}\,\pi^{(c)}_t\;
\Normal\!\Big(
\mu^{(k)}_{t\mid t-1}-\nu^{(c)}_t;\,0,\,
\Sigma^{(k)}_{t\mid t-1}+\Lambda^{(c)}_t
\Big),
\nonumber
\end{equation}
followed by normalization. In practice, gating and sparsification reduce the
$K\times C_t$ expansion.

\subsubsection{Manifold Adaptation (Lie Groups)}

On a Lie group $\mathcal{X}$ (e.g., $\SE(3)$), represent each mixture component
as a Gaussian in a consistent tangent chart $\phi(\cdot)$, apply the Euclidean
GSF updates to $\xi_t=\phi(x_t)$, and reconstruct means via $\exp(\cdot)$.
During prediction, covariances are transported through group composition using
the appropriate adjoint (first-order), yielding the manifold GSF expressions
used in the main text.

\subsubsection{One-Step GSF Summary}

Given $\{w^{(k)}_{t-1},\mu^{(k)}_{t-1},\Sigma^{(k)}_{t-1}\}_{k=1}^{K_{t-1}}$:
\begin{enumerate}[leftmargin=1.2em,itemsep=2pt]
\item \textbf{Predict:} propagate each component (linear \eqref{eq:app:pred}, or EKF/UKF/CKF per component).
\item \textbf{Update:} apply \eqref{eq:app:innov}--\eqref{eq:app:weight} (or \eqref{eq:app:mm-update} for mixture likelihoods).
\item \textbf{Control:} prune/reduce (and optionally split) to enforce $K_t\le K_{\max}$.
\end{enumerate}
On manifolds, perform all steps in the chosen chart and transport covariances
via the adjoint during prediction.

\subsection{Motion Message}
\label{appx:prediction}

We derive the GSF prediction for a single Gaussian mixand under a right-invariant
$\SE(3)$ motion model, working in the right-invariant tangent chart
$\varepsilon = \log(\mu^{-1}X)\in\mathfrak{se}(3)$.

\subsubsection{Setup}
Let $X_{t-1}\in\SE(3)$ be distributed as
$X_{t-1}=\mu\,\exp(\varepsilon)$ with
$\varepsilon\sim\mathcal{N}(0,\Sigma)\subset\mathfrak{se}(3)$.
The (right-invariant) stochastic motion model is
\begin{equation}
X_t
=
X_{t-1}\,\Delta T_t\,\exp(\nu_t),
\qquad
\nu_t \sim \mathcal{N}(0,Q_t)\subset\mathfrak{se}(3),
\nonumber
\end{equation}
with $\nu_t$ independent of $\varepsilon$.

\subsubsection{Prediction Kernel}
For a single mixand, the predicted density is
\begin{equation}
\bar p(x_t)
=
\int p(x_t\mid x_{t-1})\;
\mathcal{N}_{\mathfrak{se}(3)}
\!\Big(
\log(\mu^{-1}x_{t-1});\,0,\Sigma
\Big)\,dx_{t-1}.
\nonumber
\end{equation}

\subsubsection{First-Order Pushforward}
Write $X_{t-1}=\mu\exp(\varepsilon)$. Using the group adjoint and BCH,
\begin{equation}
\mu\exp(\varepsilon)\,\Delta T_t
=
\mu\Delta T_t\;
\exp\!\Big(
\Ad_{\Delta T_t^{-1}}\varepsilon
+
\mathcal{O}(\|\varepsilon\|^2)
\Big).
\nonumber
\end{equation}
Post-multiplying by $\exp(\nu_t)$ and applying BCH again yields
\begin{equation}
\begin{aligned}
\mu\Delta T_t\;
\exp\!\big(\Ad_{\Delta T_t^{-1}}\varepsilon\big)\;
\exp(\nu_t)
&=
\mu\Delta T_t\;
\exp\!\Big(
\Ad_{\Delta T_t^{-1}}\varepsilon
\\
&
+\nu_t +\mathcal{O}(\|\varepsilon\|^2+\|\nu_t\|^2)
\Big).
\end{aligned}
\nonumber
\end{equation}

Neglecting higher-order terms, the updated error in the right-invariant chart
at the predicted mean $\mu^-=\mu\Delta T_t$ is
\begin{equation}
\varepsilon^-
\;\approx\;
\Ad_{\Delta T_t^{-1}}\varepsilon
\;+\;
\nu_t,
\nonumber
\end{equation}
and therefore
\begin{equation}
\varepsilon^-
\sim
\mathcal{N}\!\Big(
0,\;
\Ad_{\Delta T_t^{-1}}\Sigma\Ad_{\Delta T_t^{-1}}^\top + Q_t
\Big).
\nonumber
\end{equation}
Equivalently, the pushforward of a single Gaussian mixand is (to first order)
\begin{equation}
\bar p(x_t)
\approx
\mathcal{N}_{\mathfrak{se}(3)}
\!\Big(
\log\!\big((\mu\Delta T_t)^{-1}x_t\big);\;
0,\;\Sigma^-
\Big),
\nonumber
\end{equation}
with
\begin{equation}
\Sigma^-
=
\Ad_{\Delta T_t^{-1}}\Sigma\Ad_{\Delta T_t^{-1}}^\top + Q_t.
\nonumber
\end{equation}

\subsubsection{Mixtures and Weights}
Since $\int p(x_t\mid x_{t-1})\,dx_t=1$, prediction preserves mixture weights:
if $p(x)=\sum_k w_k p_k(x)$ then $\bar p(x)=\sum_k w_k \bar p_k(x)$.

\subsubsection{Small-Increment Approximation}
If $\Delta T_t=\exp(\xi_t)$ with $\|\xi_t\|\ll 1$, then
\begin{equation}
\Ad_{\Delta T_t^{-1}}
=
I - \ad(\xi_t) + \mathcal{O}(\|\xi_t\|^2),
\nonumber
\end{equation}
and the transported covariance expands as
\begin{equation}
\begin{aligned}
\Ad_{\Delta T_t^{-1}}\Sigma\Ad_{\Delta T_t^{-1}}^\top
&=
\Sigma
-\ad(\xi_t)\Sigma
-\Sigma\,\ad(\xi_t)^\top \\
&\quad +\mathcal{O}(\|\xi_t\|^2\|\Sigma\|).
\end{aligned}
\nonumber
\end{equation}
At high update rates (small $\|\xi_t\|$) and when covariances are maintained in
the updated right-invariant chart, a common conservative approximation is
\begin{equation}
\Sigma^- \approx \Sigma + Q_t,
\nonumber
\end{equation}
which we use in our implementation in practice.

\subsection{Hypothesis Management}
\label{app:hyp_management}
In the following, we provide additional details on our hypothesis management strategy. In particular, we detail how we cluster measurements and then fuse/prune/birth hypotheses.

\mypara{Measurement Clustering.}
The global measurement message~\eqref{eq:measurement} contains $N_t K$ Gaussian
components. Many of these differ only by small perturbations around the same
physical location, induced by feature noise and PnP variability. We therefore
first reduce this redundancy by clustering the component means on
$\mathrm{SE}(3)$.

\paragraph{SE(3)-aware DBSCAN.}
Let $\mu_a$ and $\mu_b$ denote two component means on $\mathrm{SE}(3)$. We define
their Lie-algebra displacement
\begin{equation}
\small
\delta_{ab}
=
\log\!\bigl( \mu_a^{-1} \mu_b \bigr)
\in \mathfrak{se}(3) \simeq \mathbb{R}^6,
\nonumber
\end{equation}
and a group-aware distance
\begin{equation}
\small
d_\text{dbscan}(\mu_a, \mu_b)
=
\left\|\, W\,\delta_{ab}\,\right\|_2,
\nonumber
\end{equation}
where $W$ is a diagonal weighting matrix that balances translation (meters)
and rotation (radians). We run DBSCAN using this distance metric, obtaining
clusters $\mathcal{C}_c$ of component indices $j$.

\paragraph{Cluster mean and covariance.}
Each measurement component $j$ is parameterized by a mean $\mu_j$ and
covariance $\Sigma_j$ in its own tangent chart. After forming a cluster
$\mathcal{C}_c$, we compute the Riemannian (Fréchet) mean $\mu_c$ on
$\mathrm{SE}(3)$ as
\begin{equation}
\small
\mu_c
=
\arg\min_{T\in\mathrm{SE}(3)}
\sum_{j\in\mathcal{C}_c}
\alpha_j
\left\|
\log\!\bigl( T^{-1}\mu_j \bigr)
\right\|^2,
\nonumber
\end{equation}
where $\alpha_j$ are the mixture weights of the original measurement
components in Eq.~\eqref{eq:measurement}.

To aggregate uncertainty in a common tangent space, we transport each
component covariance to the tangent at $\mu_c$:
\begin{equation}
\small
\Sigma_j'
=
\mathrm{Ad}_{\mu_c^{-1}\mu_j}\,
\Sigma_j\,
\mathrm{Ad}_{\mu_c^{-1}\mu_j}^{\!\top},
\nonumber
\end{equation}
and define
\begin{equation}
\small
\xi_j
=
\log\!\bigl( \mu_c^{-1}\mu_j \bigr)
\in \mathfrak{se}(3).
\nonumber
\end{equation}
The cluster covariance must capture both
(i)~the \emph{within-component} uncertainty and
(ii)~the \emph{between-components} spread of means.
The total covariance is therefore
\begin{equation}
\small
\Sigma_c
=
\frac{1}{\beta_c}
\sum_{j\in\mathcal{C}_c}
\alpha_j
\bigl(
    \Sigma_j' + \xi_j \xi_j^\top
\bigr),
\qquad
\beta_c
=
\sum_{j\in\mathcal{C}_c}
\alpha_j.
\nonumber
\end{equation}

The merged measurement message can then be written as
\begin{equation}
\small
m_t^{\text{meas}}(x)
\;\approx\;
\sum_{c=1}^{C_t}
\bar w_c\,
\mathcal{N}_{\mathfrak{se}(3)}
\!\left(
    \log\!\bigl(\mu_c^{-1} x\bigr);
    \; 0,\Sigma_c
\right),
\label{eq:clustered-measurement}
\end{equation}
where $C_t \ll N_t K$ and $\bar w_c = \beta_c / \sum_{c'} \beta_{c'}$. In practice, we retain only the top $K{=}5$ highest-weight components whose
normalized weight exceeds a small threshold (e.g., $10^{-3}$), and discard the
remainder to maintain computational efficiency.

\mypara{Fusion and Pruning.}
We next compute the product of the motion message $m_t^{\text{mot}}$ and the
clustered measurement message~\eqref{eq:clustered-measurement}. Each existing
hypothesis (a mixand of $m_t^{\text{mot}}$) is fused with at most one
measurement cluster, since only clusters within the reachable region implied
by the motion model have non-negligible overlap.

Let $(\mu_{\text{mot}}^{(i)},\Sigma_{\text{mot}}^{(i)},w_{\text{mot}}^{(i)})$ be a motion component and $(\mu_c,\Sigma_c,\bar w_c)$ a measurement cluster. Define the relative displacement
$\delta_{ic} \;=\; \log\!\bigl((\mu_{\text{mot}}^{(i)})^{-1}\mu_c\bigr)\in\mathfrak{se}(3),
\qquad
\Sigma_c' \;=\; \mathrm{Ad}_{\Delta_{ic}^{-1}}\Sigma_c\,\mathrm{Ad}_{\Delta_{ic}^{-1}}^{\!\top},$
where $\Delta_{ic}=(\mu_{\text{mot}}^{(i)})^{-1}\mu_c$, so that $\Sigma_c'$ is expressed in the tangent at $\mu_{\text{mot}}^{(i)}$.

The fused covariance and mean (in the tangent at $\mu_{\text{mot}}^{(i)}$) are
\begin{equation}
\small
\Sigma_{i,c}^{-1}
=
\bigl(\Sigma_{\text{mot}}^{(i)}\bigr)^{-1}
+
\bigl(\Sigma_c'\bigr)^{-1},
\label{eq:fused-cov}
\end{equation}
\begin{equation}
\small
\mu_{i,c}
=
\mu_{\text{mot}}^{(i)}\,
\exp\!\Big(
\Sigma_{i,c}\,
\bigl(\Sigma_c'\bigr)^{-1}\,
\delta_{ic}
\Big).
\label{eq:fused-mean}
\end{equation}

The fused weight is proportional to the product weight times the overlap between the two Gaussians,
\begin{equation}
\small
w_{i,c}
\;\propto\;
w_{\text{mot}}^{(i)}\,\bar w_c\;
\mathcal{N}\!\Bigl(
\delta_{ic};\,0,\,
\Sigma_{\text{mot}}^{(i)}+\Sigma_c'
\Bigr).
\label{eq:fused-weight}
\end{equation}
We prune fused components whose weights $w_{i,c}$ fall below a small threshold (e.g. $10^{-3}$), indicating that the motion and measurement components are too far apart to represent a physically consistent location.

\mypara{Birth of new hypotheses.}
Some measurement clusters may have negligible overlap with all motion components
(e.g., loop closure or kidnapped robot). We model this via a ``restart'' switch
$b_t\in\{0,1\}$ with prior $p(b_t{=}1)=\epsilon$:
\[
p(x_t \mid x_{t-1},u_t,b_t)
=
\begin{cases}
p_{\text{mot}}(x_t \mid x_{t-1},u_t), & b_t=0,\\
p_{\text{birth}}(x_t \mid B_t), & b_t=1,
\end{cases}
\]
where $B_t\in\mathcal{C}^{\text{new}}_t$ indexes the measurement cluster that
initializes the new hypothesis. The birth distribution is a single Gaussian
\[
p_{\text{birth}}(x_t \mid B_t{=}c)
=
\mathcal{N}_{\mathfrak{se}(3)}\!\bigl(
\log(\mu_c^{-1}x_t);\,0,\Sigma_c
\bigr),
\]
with $p(B_t{=}c \mid b_t{=}1)\propto \bar w_c$.

Marginalizing $b_t$ yields a mixture transition kernel
\begin{equation}
\tilde p(x_t \mid x_{t-1},u_t)
=
(1-\epsilon)\,p_{\text{mot}}(x_t \mid x_{t-1},u_t)
+\epsilon\,p_{\text{birth}}(x_t).
\label{eq:motion-reinit-kernel}
\end{equation}

\section{Existing Assets, Licenses, and Terms of Use}
\label{app:asset-licenses}

This appendix documents the external assets used in the experiments and implementation. We include the datasets and benchmark assets, baseline systems, pretrained model/code assets, and reimplemented baselines used in Sections~\ref{sec:experiments} and~\ref{app:topo-bench}. The license information reflects the public project pages and repositories checked on May~5,~2026. We exclude papers cited only as related work. All listed assets are credited through the citations in the main paper or appendix. We used the assets for academic research evaluation, did not redistribute third-party datasets, model checkpoints, or baseline code as part of this work, and respected non-commercial, share-alike, and copyleft terms by using affected assets only as external evaluation baselines or ablations.

\begin{table}[h]
\centering
\caption{External datasets and benchmark assets used in the paper. ``Terms recorded'' summarizes the license or official terms of use that we found from the asset's public distribution page.}
\label{tab:dataset-licenses}
\scriptsize
\setlength{\tabcolsep}{3pt}
\renewcommand{\arraystretch}{1.08}
\begin{tabularx}{\linewidth}{@{}p{0.18\linewidth}p{0.15\linewidth}p{0.28\linewidth}X@{}}
\toprule
\textbf{Asset} & \textbf{Use in this paper} & \textbf{Terms recorded} & \textbf{Compliance note} \\
\midrule
OpenLORIS-Scene, Corridor split~\cite{shi2019openlorisscene} & Indoor relocalization benchmark & Official page describes OpenLORIS as an open dataset and requests citation; we did not find a separate named data license on the official page. The accompanying OpenLORIS-Scene tools repository is under the MIT license. & Used only for benchmark evaluation. We cite the dataset paper and do not redistribute the dataset. Project page: \href{https://lifelong-robotic-vision.github.io/dataset/scene.html}{OpenLORIS-Scene}; tools: \href{https://github.com/lifelong-robotic-vision/openloris-scene-tools}{openloris-scene-tools}. \\
ROVER, Campus scene~\cite{schmidt2025rover} & Outdoor multi-season relocalization benchmark & MIT license on the public dataset distribution page. & Used only for benchmark evaluation. We cite the dataset paper and do not redistribute the dataset. Project page: \href{https://iis-esslingen.github.io/rover/}{ROVER}; dataset page: \href{https://huggingface.co/datasets/iis-esslingen/ROVER}{Hugging Face}. \\
Topo-Bench~\cite{wang2025topo} & Perceptual-aliasing benchmark in Appendix~\ref{app:topo-bench} & Public benchmark paper states that datasets, baselines, and evaluation tools are open-sourced. The public paper distribution is under CC BY-NC-SA~4.0; any repository-specific license should govern repository files if different. & Used only for research evaluation. We cite the benchmark paper, do not redistribute benchmark assets, and follow the benchmark's non-commercial/share-alike terms where applicable. \\
Real-robot data collected for this work & Object-goal navigation and qualitative deployment experiments & New data collected by the authors; not an external asset and not released with this paper. & No third-party data license applies. Privacy and deployment considerations are discussed in Appendix~\ref{app:broader-impacts}. \\
\bottomrule
\end{tabularx}
\end{table}

\begin{table}[h]
\centering
\caption{External baseline systems, model/code assets, and their license terms.}
\label{tab:system-model-licenses}
\scriptsize
\setlength{\tabcolsep}{3pt}
\renewcommand{\arraystretch}{1.08}
\begin{tabularx}{\linewidth}{@{}p{0.17\linewidth}p{0.18\linewidth}p{0.24\linewidth}X@{}}
\toprule
\textbf{Asset} & \textbf{Use in this paper} & \textbf{License / terms recorded} & \textbf{Compliance note} \\
\midrule
ORB-SLAM3~\cite{campos2021orb} & SLAM baseline; also used inside the metric+semantic baseline & GPL-3.0. & Used as a separate external baseline for academic evaluation; we do not incorporate or redistribute ORB-SLAM3 code inside CROSS. Repository: \href{https://github.com/UZ-SLAMLab/ORB_SLAM3}{ORB-SLAM3}. \\
RTAB-Map~\cite{labbe2019rtab} & SLAM/topological baseline; also used inside the metric+semantic baseline & BSD license when OpenCV is built without nonfree modules; research-only if built with OpenCV nonfree/SURF. & Used as a separate evaluation baseline. We use the permissive/default configuration where applicable and avoid redistributing RTAB-Map binaries or databases. Project page: \href{https://introlab.github.io/rtabmap/}{RTAB-Map}. \\
MASt3R-SLAM~\cite{murai2025mast3r} & SLAM baseline & Creative Commons Attribution-NonCommercial-ShareAlike 4.0 (CC BY-NC-SA~4.0). & Used only as a non-commercial research evaluation baseline; not redistributed as part of CROSS. Repository: \href{https://github.com/rmurai0610/MASt3R-SLAM}{MASt3R-SLAM}. \\
OpenFABMAP / FAB-MAP~\cite{cummins2008fabmap2} & Classical baseline in the Topo-Bench comparison & GNU GPL v3 or later in the public openFABMAP repository. & Used only as an external benchmark baseline. We do not incorporate its code into CROSS or redistribute modified versions. Repository: \href{https://github.com/arrenglover/openfabmap}{openFABMAP}. \\
RatSLAM / OpenRatSLAM~\cite{milford2008mapping} & Classical baseline in the Topo-Bench comparison & GPL-3.0 in the public RatSLAM/OpenRatSLAM repository. & Used only as an external benchmark baseline. We do not incorporate its code into CROSS or redistribute modified versions. Repository: \href{https://github.com/davidmball/ratslam}{ratslam}. \\
BoQ / Bag-of-Queries~\cite{ali2024boq} & Visual place recognition backend used by CROSS and reimplemented topological baselines & MIT license. & Used as an off-the-shelf VPR model/code asset with attribution. Repository: \href{https://github.com/amaralibey/Bag-of-Queries}{Bag-of-Queries}. \\
XFeat~\cite{potje2024xfeat} & Local feature extraction for relative pose estimation & Apache-2.0 license. & Used as an off-the-shelf feature extractor with attribution. Repository: \href{https://github.com/verlab/accelerated_features}{accelerated\_features}. \\
LightGlue~\cite{lindenberger2023lightglue} & Local feature matching for relative pose estimation & Apache-2.0 license for LightGlue code and pretrained weights; the repository notes that SuperPoint has a different restrictive license. & Used with attribution as a matching module. We do not rely on SuperPoint-specific weights unless separately licensed. Repository: \href{https://github.com/cvg/LightGlue}{LightGlue}. \\
VGGT~\cite{wang2025vggt} & Learned relative-pose-estimation ablation & Repository license permits commercial use for the code after the July~2025 license update, but only the VGGT-1B-Commercial checkpoint is licensed for commercial usage; the original checkpoint remains non-commercial. & Used only for a research ablation and not in the final system. We do not redistribute checkpoints. Repository: \href{https://github.com/facebookresearch/vggt}{VGGT}. \\
\bottomrule
\end{tabularx}
\end{table}

\begin{table}[h]
\centering
\caption{Baselines that we reimplemented rather than importing as third-party code assets.}
\label{tab:reimplemented-baseline-licenses}
\scriptsize
\setlength{\tabcolsep}{3pt}
\renewcommand{\arraystretch}{1.08}
\begin{tabularx}{\linewidth}{@{}p{0.18\linewidth}p{0.22\linewidth}X@{}}
\toprule
\textbf{Baseline} & \textbf{Asset/license status} & \textbf{Compliance note} \\
\midrule
Greedy Matching (GM) & In-house implementation; no third-party code asset imported. & Based on the retrieval-based localization protocol described in Appendix~\ref{app:topo-baselines}; the common VPR backend is BoQ, whose MIT license is listed in Table~\ref{tab:system-model-licenses}. \\
Sequence Matching (SM)~\cite{savinov2018semi,meng2020scaling} & In-house implementation; no third-party code asset imported. & Original sequence/topological-localization papers are cited. The implementation shares only the experimental protocol and VPR backend described in Appendix~\ref{app:topo-baselines}. \\
Probabilistic Belief Update (PBU)~\cite{suomela2024placenav} & In-house implementation; no third-party code asset imported. & Original probabilistic/topometric-localization paper is cited. The implementation shares only the experimental protocol and VPR backend described in Appendix~\ref{app:topo-baselines}. \\
Appearance-Based Mapping (ABM)~\cite{labbe2013appearance} & Implemented in the RTAB-Map~\cite{labbe2019rtab} package. & Original paper is cited. \\
\bottomrule
\end{tabularx}
\end{table}

\newpage

\section{Additional Figures and Plots}

\begin{figure}[h]
    \makebox[\linewidth][l]{%
        \hspace*{0.1\linewidth}%
        \includegraphics[width=0.90\linewidth]{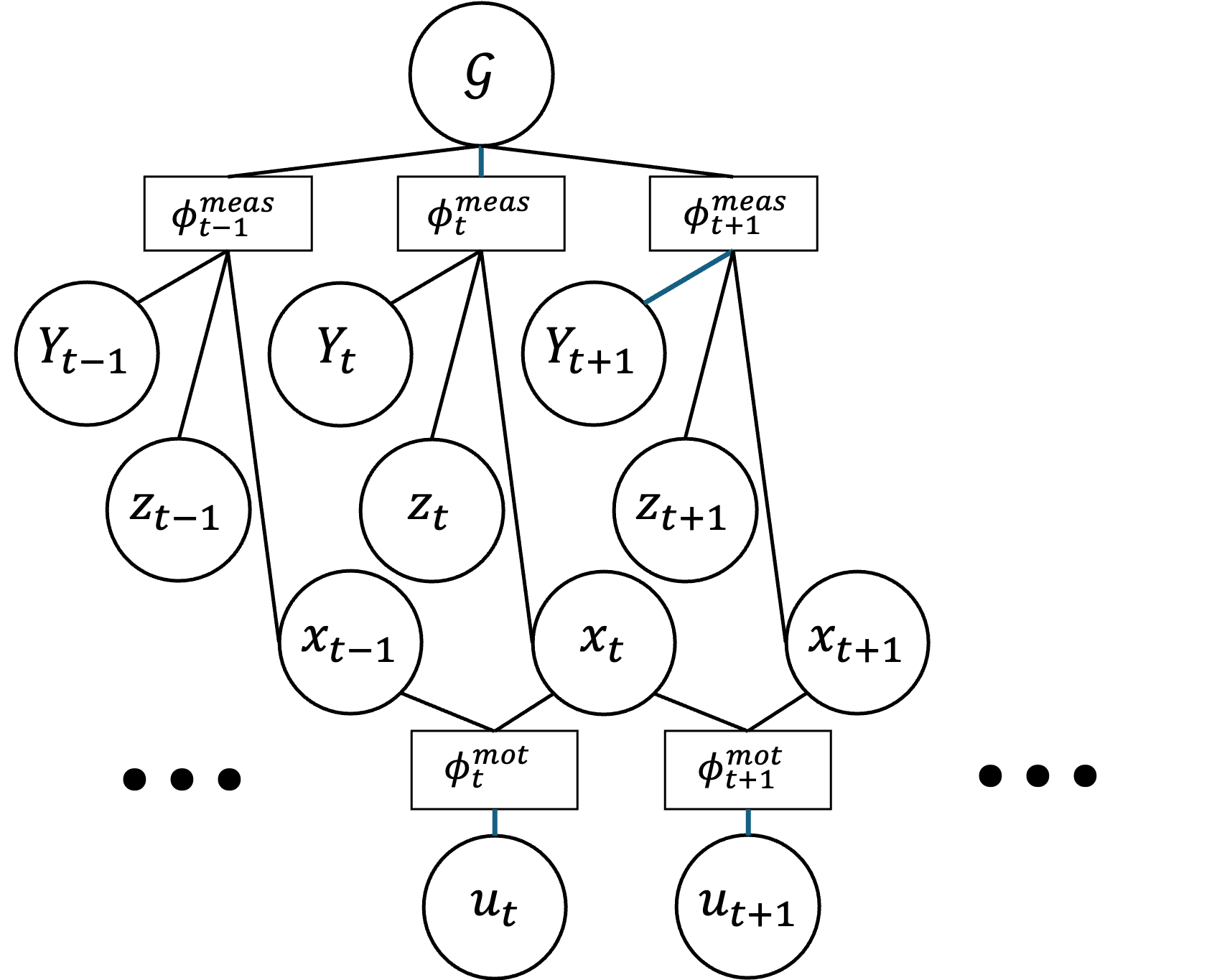}%
    }
    \caption{
    Factor-graph representation of our Gaussian mixture filtering model.
    $\phi^{\text{mot}}_t$ and $\phi^{\text{meas}}_t$ are the motion and measurement
    factors. $x_t$ is the current pose, $u_t$ the odometry input, and $z_t$ the
    RGB-D observation. $Y_t$ is a latent association variable identifying which
    keyframe explains $z_t$, and $\mathcal{G}$ is the set of stored keyframes.
    }
    \label{fig:pgm-factor-graph}
\end{figure}

\begin{figure}[t]
    \centering
    \includegraphics[width=0.99\linewidth]{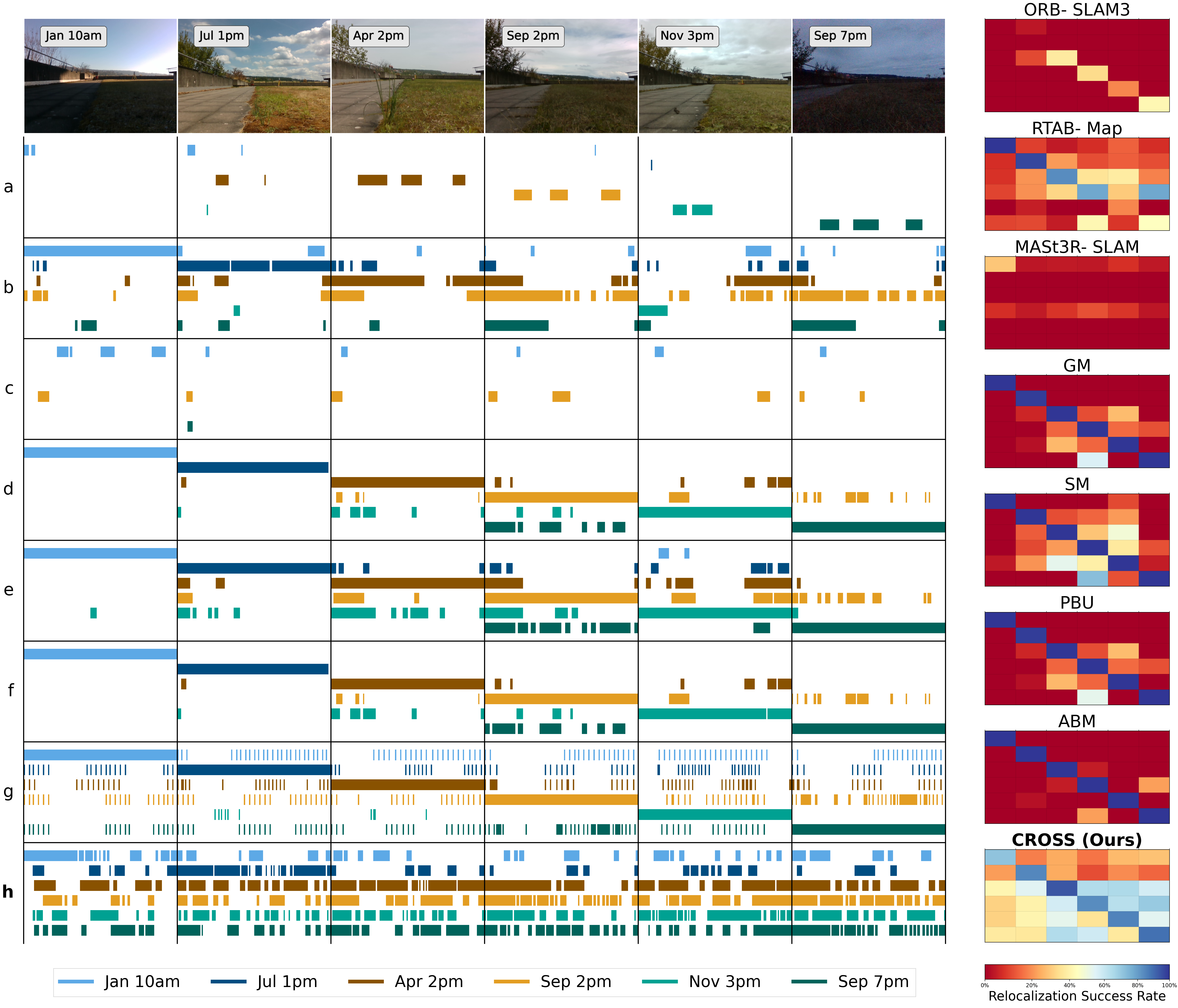}
    \caption{ Multi-session relocalization results on the Rover~\cite{schmidt2025rover} Campus scene.
\textbf{Left:} Relocalization outcomes across different locations.
Each row corresponds to a mapping trajectory (indicated by different colors), while columns
show relocalization attempts at the same physical locations captured at different
times or months, as illustrated in the top image.
Empty space indicates relocalization failed at that specific location.
The compared methods are:
(a) ORB-SLAM3~\cite{campos2021orb},
(b) RTAB-Map~\cite{labbe2019rtab},
(c) MASt3R-SLAM~\cite{murai2025mast3r},
(d) Greedy Matching (GM),
(e) Sequence Matching (SM),
(f) Probabilistic Belief Update (PBU),
(g) ABM~\cite{labbe2013appearance},
and (h) Ours.
Most baseline methods struggle under significant lighting and appearance changes,
whereas our approach consistently relocalizes despite substantial visual variation.
\textbf{Right:} A compact summary view, where each grid cell reports the
relocalization success rate for a given mapping--testing sequence pair. Additional analysis is provided in
Section~\ref{subsec:exp-appearance-change}.}
    \label{fig:campus_large_all}
\end{figure}

\begin{figure}[t]
    \centering
    \includegraphics[width=0.99\linewidth]{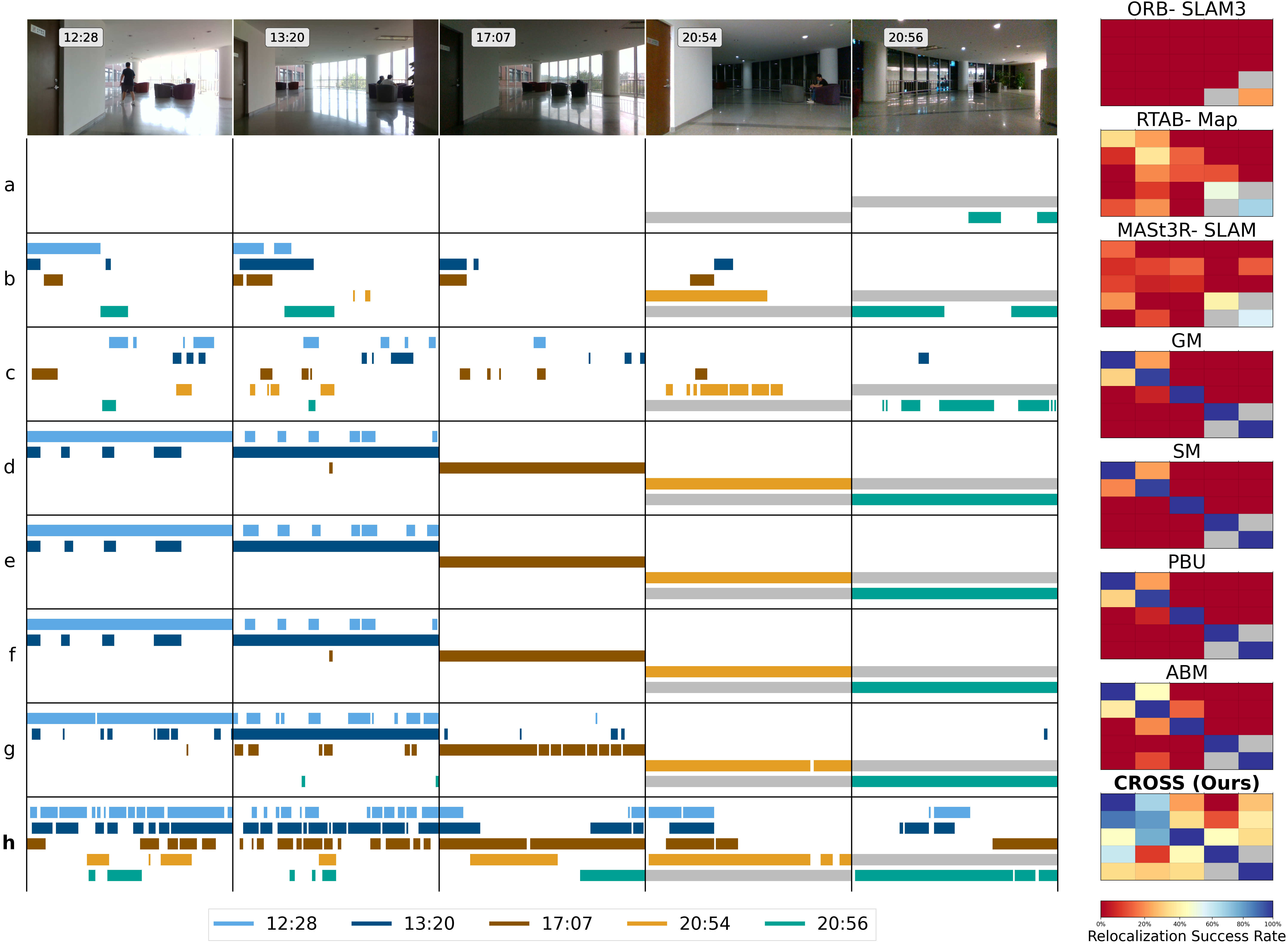}
\caption{ Multi-session relocalization results on the OpenLORIS~\cite{shi2019openlorisscene} Corridor scene.
\textbf{Left:} Relocalization outcomes across different locations.
Each row corresponds to a mapping trajectory (indicated by different colors), while columns
show relocalization attempts at the same physical locations captured at different
times or months, as illustrated in the top image.
Empty space indicates relocalization failed at that specific location. Note that the last two sequences (20:54 and 20:56) are shaded gray to indicate a lack of spatial overlap.
The compared methods are:
(a) ORB-SLAM3~\cite{campos2021orb},
(b) RTAB-Map~\cite{labbe2019rtab},
(c) MASt3R-SLAM~\cite{murai2025mast3r},
(d) Greedy Matching (GM),
(e) Sequence Matching (SM),
(f) Probabilistic Belief Update (PBU),
(g) ABM~\cite{labbe2013appearance},
and (h) Ours.
Most baseline methods struggle under significant lighting and appearance changes,
whereas our approach consistently relocalizes despite substantial visual variation.
\textbf{Right:} A compact summary view, where each grid cell reports the
relocalization success rate for a given mapping--testing sequence pair. Additional analysis is provided in
Section~\ref{subsec:exp-appearance-change}.}
    \label{fig:corridor_all}
\end{figure}

\end{document}